\documentclass{article}

\PassOptionsToPackage{numbers, compress}{natbib}
 \usepackage[preprint]{neurips_2026}

\usepackage{times}
\usepackage{multicol}
\usepackage[bookmarks=true]{hyperref}
\usepackage{amsfonts}
\usepackage{amsmath}
\usepackage{amssymb}
\usepackage[utf8]{inputenc}
\usepackage{multirow}
\usepackage[table]{xcolor}
\usepackage{xcolor}
\usepackage{hyperref}
\usepackage{array}
\usepackage{bm}
\usepackage{makecell}
\definecolor{lightlavender}{RGB}{235, 235, 255}
\usepackage{graphicx}
\usepackage{subcaption}
\usepackage{bm}
\usepackage{amssymb}

\usepackage[font=small,labelfont=bf]{caption}
\usepackage[font=footnotesize,labelfont=bf]{caption}
\usepackage{algorithm}
\usepackage{algorithmic}

\usepackage{booktabs}
\usepackage{enumitem}
\usepackage{wrapfig}
\title{Pose Distance Fields for Humanoid Robots}

\author{%
  \textbf{Yi Gu}$^{1}$\thanks{Equal contribution.} \quad
  \textbf{Yukang Gao}$^{1}$\footnotemark[1] \quad
  \textbf{Yangchen Zhou}$^{1}$\footnotemark[1] \quad
  \textbf{Xingyu Chen}$^{1}$ \quad
  \textbf{Yixiao Feng}$^{1}$
  \\[0.3em]
  \textbf{Mingle Zhao}$^{2}$ \quad
  \textbf{Yunyang Mo}$^{1}$ \quad
  \textbf{Zhaorui Wang}$^{1}$ \quad
  \textbf{Lixin Xu}$^{1}$ \quad
  \textbf{Renjing Xu}$^{1}$\thanks{Corresponding author.}
  \\[0.5em]
  $^{1}$The Hong Kong University of Science and Technology (Guangzhou)
  \\
  $^{2}$University of Macau
}

\begin{document}

\maketitle

\begin{abstract}
Pose and motion priors play a crucial role in humanoid robotics. Although such priors have been widely studied in human motion recovery (HMR) domain with a range of models, their adoption for humanoid robots remains limited, largely due to the scarcity of high-quality humanoid motion data. In this work, we introduce Pose Distance Fields for Humanoid Robots (PDF-HR), a lightweight prior that represents the robot pose distribution as a continuous and differentiable manifold. Given an arbitrary pose, PDF-HR predicts its distance to a large corpus of retargeted robot poses, yielding a smooth measure of pose plausibility that is well suited for optimization and control. PDF-HR can be integrated as a reward shaping term, a regularizer, or a standalone plausibility scorer across diverse pipelines. We evaluate PDF-HR on various humanoid tasks, including single-trajectory motion tracking, general motion tracking, style-based motion mimicry, and general motion retargeting. Experiments show that this plug-and-play prior consistently and substantially strengthens strong baselines. Code and models are publicly available on our \href{https://gaoyukang33.github.io/PDF-HR/}{\textcolor{magenta}{project page}}.


\end{abstract}    

\section{Introduction}
\label{sec:introduction}
Humanoid robots are appealing because they share human-like kinematics and can operate in the same spaces and with the same tools designed for humans~\cite{chen2025gmt, HeT-RSS-25, LiJ1-RSS-25, XueY-RSS-25}. However, generating reliable whole-body motion remains challenging. Feasible behavior must simultaneously respect joint limits, self-collision constraints, contact feasibility, and balance~\cite{pan2025agility, zhang2025hub}, while staying smooth and consistent with the task over time. In practice, even small errors in perception, modeling, or retargeting can push a solution toward physically awkward or unstable configurations~\cite{yang2025omniretarget, araujo2025retargeting}. A natural way to improve robustness is to incorporate pose or motion priors that encode which configurations are plausible for a given robot, and to use these priors to regularize estimation and control.

In the human motion recovery (HMR) literature, pose and motion priors~\cite{bogo2016keep, kanazawa2018end, SMPL-X:2019, lu2025dposer, he2024nrdf} have been studied extensively, supported by abundant, high-quality datasets~\cite{AMASS:ICCV:2019} and powerful generative models. These priors help resolve ambiguity from partial observations and improve robustness to noise and occlusion. In contrast, comparatively fewer works explore similarly general and reusable priors for humanoid robots. This gap is not merely historical. Collecting large-scale, high-quality humanoid robot motion data is expensive, and directly transferring priors from human bodies to robot morphologies is nontrivial due to differences in joint limits, kinematic structure, actuation capabilities, and contact dynamics. As a result, many robotics pipelines still rely on task-specific constraints, hand-tuned regularizers, or priors that are tightly coupled to a particular controller or dataset, limiting generalization across tasks~\cite{peng2018deepmimic, peng2021amp, luo2023perpetual}.

This paper focuses on pose priors and proposes a simple but effective alternative. Instead of modeling the full generative distribution of humanoid motion, we learn a continuous distance-to-data function in pose space. Concretely, we introduce \textbf{P}ose \textbf{D}istance \textbf{F}ields for \textbf{H}umanoid \textbf{R}obots (PDF-HR), a lightweight prior represented by an MLP that maps an arbitrary robot pose to a scalar distance to the nearest pose in a large corpus of retargeted robot poses. To obtain a well-behaved distance field, we carefully design the training distribution to cover the pose space while emphasizing high-quality near-manifold samples, and we use cross-validation to select reliable positive samples. The resulting function is continuous and provides meaningful gradients both near and far from the data manifold. Intuitively, PDF-HR acts like a ``pose plausibility scorer'': poses near the dataset manifold receive low distance, while out-of-distribution or physically awkward configurations receive higher distance. This distance-field formulation offers several practical advantages. First, it supports arbitrary query poses and provides a smooth signal even when the pose is far from the data manifold, which is valuable in early-stage optimization and when tracking difficult motions. Second, it is modular and reusable: PDF-HR can serve as an independent pose-quality objective without requiring additional data collection, and it can be applied across tasks and policies without retraining~\cite{mu2025smp}. Third, the model is compact and efficient, making it easy to integrate into existing systems.

We demonstrate the breadth of this approach across representative humanoid tasks, including single-trajectory motion tracking, general motion tracking, style-based motion mimicry, and motion retargeting. For tracking, we incorporate PDF-HR as a general-purpose reward shaping term that encourages exploration and optimization within the near manifold region. Empirically, the pose reward accelerates convergence for RL tracking baselines and improves robustness on challenging motions. For retargeting, PDF-HR acts as a principled regularizer that improves motion quality. Across settings, PDF-HR consistently strengthens strong baselines while remaining lightweight and easy to deploy. Our main contributions can be summarized as follows:
\begin{itemize}[leftmargin=*]
\item PDF-HR: a continuous and differentiable pose distance field for humanoid robots learned from a large corpus of retargeted robot poses.

\item Plug-and-play integration: simple mechanisms to incorporate PDF-HR as a prior across diverse humanoid applications, including motion tracking and motion retargeting.

\item Empirical validation: experiments on multiple humanoid tasks demonstrating consistent improvements in performance and robustness over strong baselines.
\end{itemize}

\section{Related Work}
\label{sec:related_works}
\noindent
\textbf{Human Motion Priors.}
Data-driven human pose and motion priors are fundamental for human motion recovery~\cite{bogo2016keep,lu2023dposer,SMPL-X:2019,rempe2021humor}. Early works mainly focused on learning explicit constraints, such as joint limits, to avoid implausible poses~\cite{akhter2015pose}. Subsequent approaches leveraged generative models, including Gaussian Mixture Models (GMMs)~\cite{bogo2016keep}, Variational Autoencoders (VAEs)~\cite{dwivedi2024tokenhmr,SMPL-X:2019}, Generative Adversarial Networks (GANs)~\cite{georgakis2020hierarchical, kanazawa2018end} and diffusion models~\cite{lu2023dposer, lu2025dposer, zhang2024rohm, shan2023diffusion} to capture the statistical distribution of natural human articulation. More recently, implicit representations like Pose-NDF~\cite{tiwari2022pose} has been proposed to model the manifold of valid poses as the zero-level set of a learned neural distance field. However, Pose-NDF relies on Euclidean gradient descent followed by re-projection onto $\mathrm{SO}(3)$ during optimization, which can limit convergence speed. In contrast, NRDF~\cite{he2024nrdf} enforces strict geometric consistency by leveraging Riemannian gradient descent, ensuring that optimization iterates always remain on the manifold. Building upon this geometric metric, NRMF~\cite{yu2025geometric} extends the representation to the temporal domain, modeling distance fields over pose, velocity, and acceleration to ensure higher-order motion continuity and dynamic feasibility. In this work, we extend this paradigm of implicit manifold representation from humans to humanoids. Specifically, we learn PDF-HR to capture the valid configuration space of the robot. We demonstrate that this learned prior can significantly enhance performance in various downstream tasks.

\noindent
\textbf{Physics-based Motion Imitation.} 
Data-driven imitation learning has emerged as a cornerstone for acquiring humanoid skills~\cite{coros2010generalized,yin2007simbicon}. DeepMimic~\cite{peng2018deepmimic} pioneered this direction by tracking reference motions to produce physically feasible behaviors. However, such tracking-based methods typically require the controller to rigidly mimic targets frame-by-frame, limiting the flexibility to adapt to new tasks~\cite{liu2017learning}. Furthermore, they often necessitate labor-intensive reward tuning, a challenge partially addressed by introducing Adversarial Differential Discriminator (ADD)~\cite{zhang2025physics}. To overcome the constraints of rigid tracking and paired data alignment, distribution-matching approaches such as Generative Adversarial Imitation Learning (GAIL)~\cite{ho2016generative} and Adversarial Motion Priors (AMP)~\cite{peng2021amp} have been widely adopted. By learning flexible motion priors from datasets, these methods provide a task-agnostic measure of naturalness, allowing policies to produce life-like behaviors across diverse tasks. This paradigm was further expanded by ASE~\cite{peng2022ase} to amplify motion diversity and repurpose skills hierarchically.
The closest method to ours is SMP~\cite{mu2025smp}, which leverages pre-trained motion diffusion models with score distillation sampling (SDS)~\cite{PooleJBM23} to extract reusable, task-agnostic motion priors. Our work follows a similar \textbf{modular} and \textbf{reusable} philosophy, but tackles the problem from a different perspective: pose rather than motion. We also show that our prior improves style-conditioned tracking, making it theoretically complementary to SMP.

\noindent
\textbf{Motion Retargeting.} Retargeting motions from human demonstrations to humanoid robots introduces unique challenges, particularly the strict requirements for physical plausibility and consistent interactions with the environment. Geometric approaches~\cite{he2024omnih2o,he2025asap,luo2023perpetual,araujo2025retargeting} typically minimize keypoint errors via trajectory optimization or inverse kinematics. However, when physical constraints are weak or absent, they can violate kinematic and contact limits, leading to artifacts such as foot skating and interpenetration. To improve retargeting fidelity, subsequent works~\cite{videomimic, lee2025phuma} introduced soft interaction penalties, and OmniRetarget~\cite{yang2025omniretarget} further unifies these objectives by minimizing the interaction mesh~\cite{nakaoka2012interaction} Laplacian deformation energy under explicit hard constraints, yielding more robust physical consistency. Complementary to these optimization-based formulations, SPIDER~\cite{pan2025spider} leverages large-scale physics-based sampling in a simulator with curriculum-style virtual contact guidance to resolve contact ambiguity and recover feasible interactions. To the best of our knowledge, PDF-HR is the first learned prior that can be introduced as a regularizer into the optimization loop of existing pipelines, encouraging solutions to remain within the valid configuration space of the humanoid and thereby improving both motion naturalness and physical feasibility.

\section{Background: Riemannian Geometry}
\label{sec:background}

We first establish the preliminaries needed to formalize our pose distance fields. Following recent works~\cite{birdal2019probabilistic, birdal2018bayesian, chen2022projective, he2024nrdf,yu2025geometric}, we consider a smooth $m$-dimensional Riemannian manifold $\mathcal{M}$, smoothly embedded in a higher-dimensional ambient space $\mathcal{X}$ (typically $\mathbb{R}^n$) via an embedding $\iota: \mathcal{M} \hookrightarrow \mathcal{X}$. The manifold $\mathcal{M}$ is endowed with a Riemannian metric $\mathbf{G} \triangleq (\mathbf{G}_{\mathbf{x}})_{\mathbf{x}\in\mathcal{M}}$, where each $\mathbf{G}_{\mathbf{x}}$ defines an inner product on the tangent space $T_{\mathbf{x}}\mathcal{M}$ that varies smoothly with $\mathbf{x}$, forming a smooth Riemannian manifold $(\mathcal{M}, \mathbf{G})$. In composite systems, especially for articulations, we exploit the structure of product manifolds~\cite{zhang2021product, woodward2024product}. Given $K$ component manifolds $\mathcal{M}_1, \dots, \mathcal{M}_K$, their Cartesian product $\mathcal{M}_{1:K} = \mathcal{M}_1 \times \dots \times \mathcal{M}_K$ is a smooth manifold of dimension $\sum_{i=1}^{K}\dim(\mathcal{M}_i)$. Moreover, equipping it with the canonical product metric yields a Riemannian manifold. When all $K$ components are identical (i.e., $\mathcal{M}_i \equiv \mathcal{M}_j$), we denote the configuration space as the power manifold $\mathcal{M}^K$, which has dimension $Km$ and inherits a natural product Riemannian structure~\cite{zhang2021product, woodward2024product, he2024nrdf,yu2025geometric}. 

In the context of HMR and humanoid robotics, the kinematic state of each joint is commonly formalized as a 3D rotation in the Special Orthogonal group $\mathrm{SO}(3)$. This Lie group is defined as:
\begin{equation}
\mathrm{SO}(3) = \left\{ \mathbf{R} \in \mathbb{R}^{3\times 3} \ \middle|\ \mathbf{R}^\top \mathbf{R}=\mathbf{I},\ \det(\mathbf{R})=1 \right\}.
\end{equation}
Viewing $\mathrm{SO}(3)$ as a submanifold embedded in $\mathbb{R}^{9}$ (identified with $\mathbb{R}^{3\times 3}$), the full configuration space of a skeleton with $K$ joints can be constructed as the power manifold of Lie groups:
\begin{equation}
\mathcal{M}_H \triangleq \prod_{k=1}^{K} \mathrm{SO}(3)_k \;=\; \mathrm{SO}(3)^K.
\end{equation}
Since $\mathrm{SO}(3)$ is a Lie group, it admits well-defined tangent spaces and the exponential/logarithmic maps, which allow us to perform local (linear) computations in the tangent space while respecting the underlying manifold geometry.

\begin{figure*}[!t] 
\centering
\includegraphics[width=0.99\linewidth]{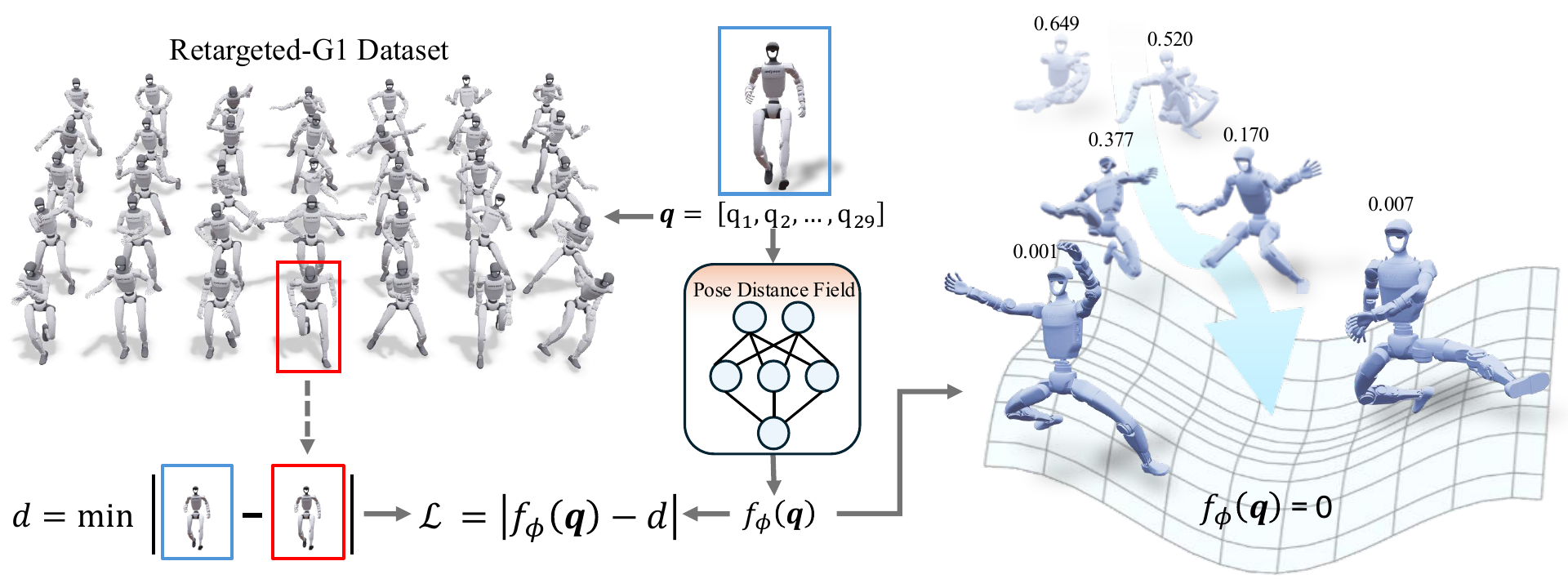}
\caption{We present \textbf{PDF-HR}, which learns the manifold of plausible G1 poses as a zero-level set. \textit{Left}: The $f_\phi$ is trained to approximate the unsigned pose distance field. Given a query pose $\mathbf{q}$, we compute its distance $d$ to the nearest dataset sample and optimize the network to regress this value. \textit{Right}: The learned prior provides quantitative scores for arbitrary poses, where a larger predicted value $f_\phi(\mathbf{q})$ indicates a significant deviation from the manifold, corresponding to an unnatural pose. This learned prior effectively benefits downstream tasks such as motion tracking and motion retargeting.
}
\label{fig:pipeline} 
\end{figure*}

\section{Methods}
\label{sec:methods}
In contemporary humanoids, the articulated body is commonly modeled as a kinematic tree of 1-DoF revolute joints, which enables direct actuation, high structural stiffness, and well-conditioned dynamics. This design choice facilitates efficient whole-body control, optimization, and learning. In this work, we adopt the Unitree G1 humanoid as our primary platform. For a 1-DoF joint, the geodesic metric simplifies to the Euclidean distance. Let $q_i, q_i' \in \mathbb{R}$ denote two configurations of the same joint. The corresponding rotations can be expressed as:
\begin{equation}
    \mathbf{R}_i = \exp([\mathbf{k}]q_i), \|\mathbf{k}\|=1,
\end{equation}
where $\mathbf{k}\in \mathbb{R}^3$ is the fixed rotation axis and $[\mathbf{k}] \in \mathfrak{so}(3)$ denotes its associated skew-symmetric matrix. The relative rotation $\mathbf{R}_{rel} = \mathbf{R}_i^\top \mathbf{R}_{i}'$ can be calculated by:
\begin{equation}
\label{eq:Rot_rel}
    \mathbf{R}_{rel} = \exp(-[\mathbf{k}]{q}_i) \exp([\mathbf{k}]{q}_i') = \exp\!\left([\mathbf{k}]({q}_i' - {q}_i)\right).
\end{equation}

The geodesic distance induced by the canonical bi-invariant Riemannian metric on $\mathrm{SO}(3)$ is:
\begin{equation}\label{eq:geo_def}
d_{\mathrm{geo}}(\mathbf{R}_1,\mathbf{R}_2)
=
\left\|\big(\log(\mathbf{R}_1^\top\mathbf{R}_2)\big)^{\vee}\right\|_2,
\end{equation}
where $(\cdot)^{\vee}:\mathfrak{so}(3)\rightarrow\mathbb{R}^3$ is the inverse of the skew operator, and $\|\cdot\|_2$ is the Euclidean norm.

Substituting Eq.~\eqref{eq:Rot_rel} into Eq.~\eqref{eq:geo_def}, the geodesic metric strictly reduces to the $L_1$ distance:
\begin{equation}
    d(\mathbf{R}_i, \mathbf{R}_i') =|{q}_i - {q}_i'|,
\end{equation}
assuming $q_i$ is represented within its joint limits (i.e., no angle wrap-around).

\noindent
\textbf{Pose Manifold and Distance Field Learning.} We hypothesize that physically plausible robot poses lie on a low-dimensional manifold $\mathcal{M_{HR}}$ embedded in the high-dimensional configuration space $\mathcal{Q} \subseteq \mathbb{R}^{N_J}$. $\mathcal{Q}$ is a proper subset of $\mathbb{R}^{N_J}$ due to per-joint bound constraints specified by the URDF. The instantaneous configuration of the Unitree G1 robot is represented as a vector $\mathbf{q}$ describing the articulation of its $N_J=29$ revolute joints:
\begin{equation}
     \mathbf{q} = [q_1, q_2, \dots, q_{N_J}]^\top \in \mathcal{Q},
    \label{eq:state_def}
\end{equation}
where $q_i$ is a scalar joint angle. Endowed with the $L_1$ product metric, the distance between two configurations is:
\begin{equation}
     d_\mathcal{M_{HR}}(\mathbf{q}, \mathbf{q'}) = \sum_{i=1}^{N_J} d(\mathbf{R}_i, \mathbf{R}_i').
\end{equation}

Instead of explicitly parameterizing this manifold, we model $\mathcal{M_{HR}}$ implicitly as the zero-level set of a continuous unsigned distance function $f_\phi: \mathcal{Q} \rightarrow \mathbb{R}^{+}$:

\begin{equation}
    \mathcal{M_{HR}} = \{\, \mathbf{q} \in \mathcal{Q} \mid f_\phi(\mathbf{q}) = 0 \,\},
\end{equation}
where $f_\phi(\mathbf{q})$ represents the unsigned geodesic distance to the closest plausible pose on the manifold. 

Training an accurate distance field $f_\phi$ requires a large number of samples, including both positive (on-manifold, i.e., zero-distance) and negative (off-manifold, i.e., non-zero-distance) poses. Leveraging recent advances in retargeting, we construct a large-scale positive set $\mathcal{D}_p$ using the PHUMA~\cite{lee2025phuma}, LaFAN1, and AMASS~\cite{AMASS:ICCV:2019} retargeted datasets. We further perform cross-validation to filter unreliable positives. Despite having abundant on-manifold samples, the performance of $f_\phi$ depends critically on the statistical coverage of the training distribution and, in particular, on incorporating informative off-manifold negatives $\mathcal{D}_n$. Motivated by NRDF~\cite{he2024nrdf}, we therefore adopt a hybrid sampling strategy to construct $\mathcal{D}_n$. Details of the hybrid sampling procedure and the cross-validation protocol are provided in the Appendix A.


With the resulting dataset $\mathcal{D}=\mathcal{D}_p\cup \mathcal{D}_n$ containing $N$ samples, we learn a continuous unsigned distance field via supervised regression. Specifically, $f_\phi$ is trained to approximate the minimum geodesic distance from a query pose $\mathbf{q}$ to the valid pose manifold $\mathcal{M}_{\mathcal{HR}}$, which is approximated by the distance to the nearest sample in the dense dataset $\mathcal{D}_p$. The optimal parameters are obtained by minimizing the objective function:
\begin{equation}
\phi^\star
= \arg\min_{\phi}\sum_{i=1}^{N}
\left\|
f_\phi(\mathbf{q}_i)
- \min_{\mathbf{q}'\in\mathcal{D}_p} d_\mathcal{M_{HR}}(\mathbf{q}_i,\mathbf{q}')
\right\|_{1}.
\end{equation}
Once trained, $f_\phi$ serves as a continuous and differentiable pose prior that explicitly captures the geometry of the valid pose manifold. PDF-HR is also lightweight, with only 0.42M parameters, and runs at 635.5 FPS on an RTX 3090 with batch size 1, making it well suited for RL training with frequent simulator interactions. Figure~\ref{fig:pipeline} illustrates the full training pipeline.

For 1-DoF joints, the Riemannian gradient coincides with the Euclidean gradient (proven in Appendix C). Leveraging this differentiability, the projection of an arbitrary query pose $\mathbf{q}$ onto the manifold $\mathcal{M}$ can be formulated as an iterative gradient-based update:
\begin{equation}
\label{eq:gradient}
\mathbf{q}_{k+1} = \mathcal{P} \left( \mathbf{q}_k - \alpha \cdot f_\phi(\mathbf{q}_k) \cdot \frac{\nabla_{\mathbf{q}} f_\phi(\mathbf{q}_k)}{\|\nabla_{\mathbf{q}} f_\phi(\mathbf{q}_k)\|} \right)
\end{equation}
where $\alpha$ is the step size, $-\nabla_{\mathbf{q}} f_\phi(\mathbf{q}_k)/\|\nabla_{\mathbf{q}} f_\phi(\mathbf{q}_k)\|$ denotes the gradient direction towards the manifold. The operator $\mathcal{P}$ enforces physical joint limits, ensuring the updated pose remains within the valid configuration space.

\section{Applications}
\label{sec:applications}
\subsection{Reinforcement Learning based Motion Tracking}
\noindent
Reinforcement learning (RL) is a standard approach for motion tracking. Modern RL-based tracking methods generally fall into two categories: strict tracking~\cite{peng2018deepmimic, zhang2025physics} and stylized tracking~\cite{peng2021amp, mu2025smp}. In both cases, the reward can be simplified as:
\begin{equation}
    r_t = w^{G} r_t^{G} + w^{T} r_{t}^{T}.
\end{equation}
Here, $r_t^{G}$ is the task reward, which defines high-level goals for the character to satisfy (e.g., walking in a target direction or punching a target). $r_{t}^{T}$ denotes the tracking reward. For strict tracking, it encourages the character to imitate the reference motion as closely as possible, typically by minimizing the pose discrepancy between the simulated character and the target poses from the reference trajectory. In contrast, for stylized tracking, $r_{t}^{T}$ encourages similarity in motion style and is not directly tied to pose error. $w^{T}$ and $w^{G}$ are the corresponding reward weights.

Although RL-based tracking methods can successfully follow most motions, they typically require long training times to explore a sufficiently diverse set of states. To improve training efficiency, we introduce a pose-prior reward that encourages the policy to explore near the learned pose manifold. Specifically, we first define a pose score $s$ as
\begin{equation}
    \label{eq:pose-prior score}
    s(\mathbf{q}_t) = \mathrm{clip}\!\left(\frac{f_\phi(\mathbf{q}_t)-d_{good}}{d_{bad}-d_{good}}, 0, 1\right)
\end{equation}
where $d_{bad}$ is a scalar hyperparameter. In all experiments, we set $d_{bad}=0.4$ (i.e., if the predicted pose distance $f_\phi(\mathbf{q}_t)$ exceeds $0.4$, we treat the pose as abnormal). $d_{good}$ is computed from the reference motion as the maximum $f_\phi(\mathbf{q}_t)$ observed along the trajectory. We clip $s(\mathbf{q}_t)$ to $[0, 1]$ for training stability. We then define the pose-prior reward $r_t^P$ as:
\begin{equation}
    \label{eq:pose-prior reward}
    r_t^P = e^{-\alpha^P s(\mathbf{q}_t)}.
\end{equation}
Here $\alpha^P$ controls the scale of the pose-prior term. Thus, our full reward can be formulated as:
\begin{equation}
    \label{eq:full reward}
    r_t = w^{G} r_t^{G} + w^{T} r_{t}^{T} + w^{P} r_t^{P},
\end{equation}
where $w^P$ is the weight of the pose reward $r_t^P$.




\subsection{Motion Retargeting}
\noindent
Physical humanoids only approximate human morphology, with significant differences in shape, proportion, and degrees of freedom; this is also called the embodiment gap. A common way to bridge this gap is to retarget human motion data to a humanoid embodiment and then train RL policies to imitate the resulting reference trajectories. Existing works~\cite{yang2025omniretarget} have shown that high-quality retargeting algorithms can substantially alleviate extensive reward engineering. For clarity, we adopt GMR~\cite{araujo2025retargeting} as a representative retargeting method and illustrate how to incorporate our pose prior. GMR~\cite{araujo2025retargeting} performs retargeting by solving an inverse kinematics (IK) problem at each frame.
At each timestep $t$, GMR finds the robot configuration $\mathbf{q}_t$ that matches the source keypoint positions and orientations via the following optimization program:
\begin{equation} \label{eq:gmr_qp}
    \begin{aligned}
        \mathbf{q}_t^\star &= \arg\min_{\mathbf{q}_t} \sum_i \left\|f_i^p(\mathbf{q}_t) - \mathbf{p}^\text{source}_{t, i}\right\|_2 + 
        \left\|f_i^\theta(\mathbf{q}_t) - \mathbf{\theta}^\text{source}_{t, i}\right\|_2, 
      \text{s.t. } \mathbf{q}_{\min} \leq \mathbf{q}_t \leq \mathbf{q}_{\max},
    \end{aligned}
\end{equation}
where $f_i^p$ and $f_i^\theta$ are the robot forward kinematics for the $i$-th keypoint's position and orientation, respectively. Leveraging the Mink~\cite{Zakka_Mink_Python_inverse_2025} library, GMR solves this program in a Sequential Quadratic Programming fashion.

A key limitation of GMR is the need for tedious parameter tuning, especially careful per-joint weighting, to avoid optimization artifacts. Since our pose distance field is continuous and differentiable, we can simply add an extra prior term to Eq.~\eqref{eq:gmr_qp}. The resulting objective becomes:
\begin{equation}
    \begin{aligned}
        \mathbf{q}_t^\star &= \arg \min_{\mathbf{q}_t} \sum_i \left\|f_i^p(\mathbf{q}_t) - \mathbf{p}^\text{source}_{t, i}\right\|_2 + \left\|f_i^\theta(\mathbf{q}_t) - \mathbf{\theta}^\text{source}_{t, i}\right\|_2
        + \left\|f_\phi(\mathbf{q}_t)\right\|_2\\
      & \text{s.t. } \mathbf{q}_{\min} \leq \mathbf{q}_t \leq \mathbf{q}_{\max},
    \end{aligned}
\end{equation}

\section{Experiments}
\label{sec:experiments}

To validate the effectiveness of our method, we integrate it into four humanoid tasks: single-trajectory motion tracking, general motion tracking, style-based motion mimicry, and general motion retargeting. Comprehensive details regarding the implementation and experimental configurations are provided in the Appendix D.
\begin{figure*}[!t] 
\centering
\includegraphics[width=0.99\linewidth]{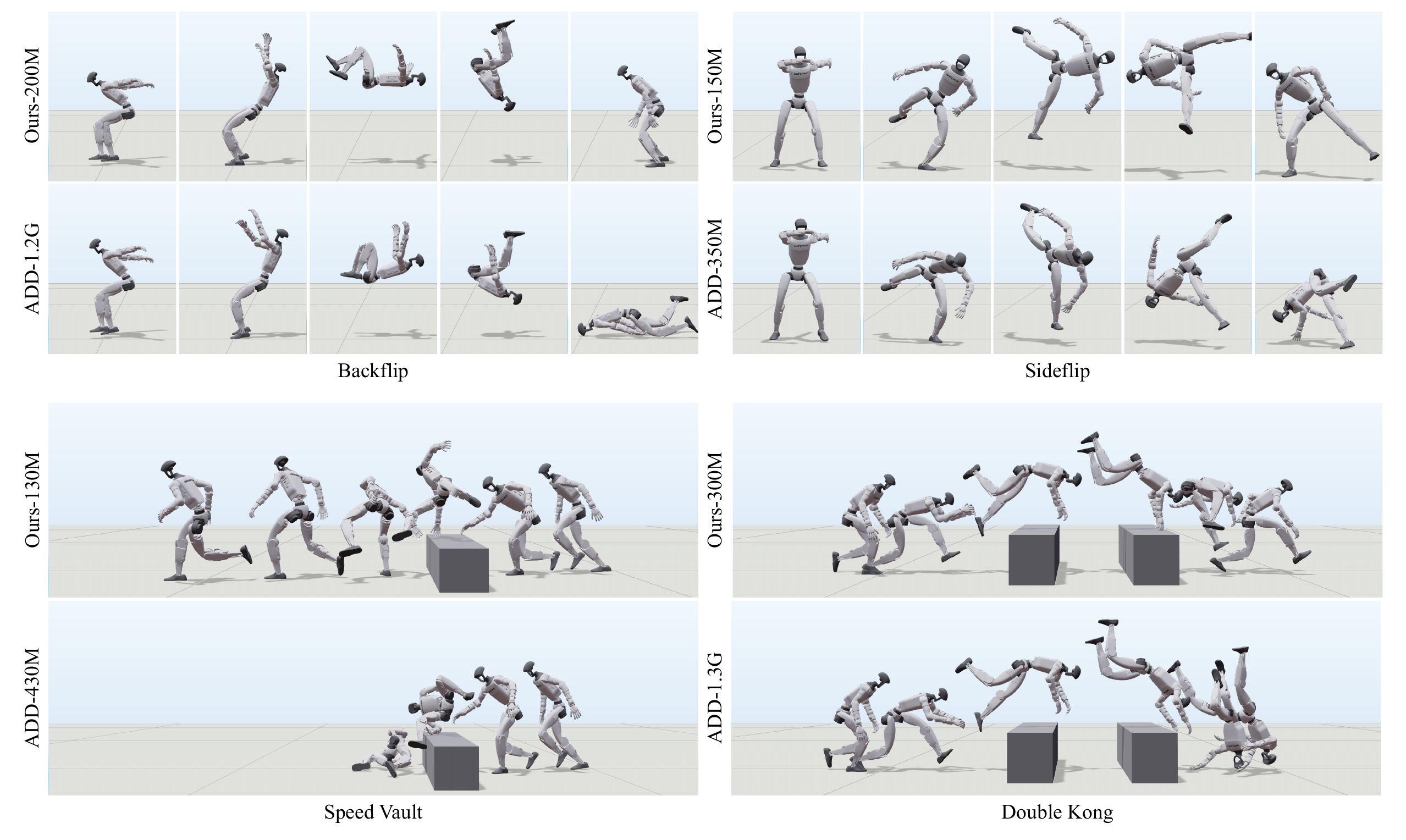}
\caption{
Visual comparisons of motion tracking performance on dynamic skills between our method and ADD~\cite{zhang2025physics}. The number of training samples is annotated on the left of each strip. Our method successfully masters these complex skills with remarkably fewer samples, whereas the baseline frequently suffers from falls or collisions even after extensive training.
}
\label{fig:ADD_single_clip} 
\end{figure*}


\noindent
\textbf{Metrics.} We evaluate our method against baselines in terms of sample efficiency and motion tracking error. All policies are trained with 4096 parallel environments and evaluated on 4096 test episodes every 100 training iterations. We define the motion tracking success rate as the ratio of continuous tracking duration to the total episode length. The sample efficiency is measured by the number of training samples required to first reach $80\%$ success rate. The position error $E^\mathrm{pos}$ and rotation error $E^\mathrm{rot}$ are defined as:
\begin{equation}
    E^\mathrm{pos} = \frac{1}{N+1} \left( \sum_{j \in \mathrm{joints}}e_j^{\mathrm{pos}} + e_{\mathrm{root}}^\mathrm{pos} \right),\ 
    E^\mathrm{rot} = \frac{1}{N+1} \left( \sum_{j \in \mathrm{joints}}e_j^{\mathrm{rot}} + e_{\mathrm{root}}^\mathrm{rot} \right),
\end{equation}
where $e_{j}^\mathrm{pos}$, $e_{\mathrm{root}}^\mathrm{pos}$, $e_{j}^\mathrm{rot}$, and $e_{\mathrm{root}}^\mathrm{rot}$ represent the time-averaged $j$-th joint position error, root position error, $j$-th joint rotation error, and root rotation error, respectively. Joint errors are computed in the robot’s local coordinate frame, while root errors are computed in the global coordinate frame. For general motion tracking, we additionally report the detailed errors (i.e., joint position error $\bar{e}^\mathrm{pos}_{joint} = \left( \sum_{j \in \mathrm{joints}}e_j^{\mathrm{pos}}\right)/N$, root position error $e_{\mathrm{root}}^\mathrm{pos}$, joint rotation error $\bar{e}^\mathrm{rot}_{joint}=\left( \sum_{j \in \mathrm{joints}}e_j^{\mathrm{rot}} \right)/N$ and root rotation error $e_{\mathrm{root}}^\mathrm{rot}$) for a more fine-grained comparison.

\subsection{Single-trajectory Motion Tracking}
\label{subsec:single_traj_tracking_exp}

\noindent
\textbf{Experiment settings.} We compare our method with ADD~\cite{zhang2025physics} in the single-trajectory tracking task using nine motion clips: three easy motions (Walk, Run, Jump), four hard motions (Spinkick, Cartwheel, Backflip, Sideflip), and two parkour motions (Double Kong, Speed Vault). The parkour skills are particularly challenging, as they require high tracking accuracy to reproduce intricate contacts with the environment. For a fair comparison, all policies are trained with 800 million samples with the same hyperparameters.

\begin{table*}[!ht]
\scriptsize
\setlength{\tabcolsep}{4pt}
\renewcommand{\arraystretch}{1.5} 
\centering
\caption{Comparison of sample efficiency and motion-tracking performance between ADD~\cite{zhang2025physics} and our method on the single-trajectory motion-tracking task. Metrics are reported as mean $\pm$ 1 std over three independently trained models with different random initializations. A motion is marked as \textbf{Failed} if all three trials fail to converge within $800\text{M}$ training samples.
}
\label{tab:ADD_singleclip_table}
\begin{tabular}{|c|c|c|c|c|c|c|c|}
\hline

\textbf{Skill} & \textbf{Length} & \multicolumn{2}{c|}{\textbf{Samples ($SR \geq 80\%$ ) [M]} } & \multicolumn{2}{c|}{\textbf{Position Error [m]}} & \multicolumn{2}{c|}{\textbf{Rotation Error [rad]}} \\ \hline

               &                 & \textbf{ADD} & \textbf{Ours} & \textbf{ADD} & \textbf{Ours} & \textbf{ADD} & \textbf{Ours} \\ 
\hline
Walk    &      1.03       & $70.036^{\pm 6.179}$ & \cellcolor{blue!10}\({56.929^{\pm 6.179}}\) & \cellcolor{blue!10}\({0.008^{\pm 0.000}}\) & $0.009 ^{\pm 0.000}$ & \cellcolor{blue!10}\({0.038^{\pm 0.001}}\) & $0.042^{\pm 0.002}$ \\ 
\hline
Run    &      0.80       & $113.727^{\pm 6.179}$ & \cellcolor{blue!10}\({70.036^{\pm 6.179}}\) & \cellcolor{blue!10}\({0.015^{\pm 0.000}}\) & $0.017^{\pm 0.000}$ & \cellcolor{blue!10}\({0.063^{\pm 0.001}}\) & $0.070^{\pm 0.001}$\\ 
\hline
Jump    &      1.75       & $222.953^{\pm 10.702}$ & \cellcolor{blue!10}\({188.001^{\pm 16.348}}\) & \cellcolor{blue!10}\({0.022^{\pm 0.001}}\) & $0.024^{\pm 0.001}$ & $0.108^{\pm 0.004}$ & \cellcolor{blue!10}\({0.103^{\pm 0.003}}\) \\ 
\hline
Spinkick    &      1.28       & $135.572^{\pm 6.179}$ & \cellcolor{blue!10}\({91.881^{\pm 0.000}}\) & \cellcolor{blue!10}\({0.031^{\pm 0.000}}\) & $0.032^{\pm 0.001}$ & \cellcolor{blue!10}\({0.150^{\pm 0.002}}\) & $0.161^{\pm 0.004}$ \\ 
\hline
Cartwheel    &      2.72       & $183.632^{\pm 0.000}$ & \cellcolor{blue!10}\({161.787^{\pm 12.358}}\) & $0.031^{\pm 0.000}$ & \cellcolor{blue!10}\({0.028^{\pm 0.003}}\) & $0.129^{\pm 0.007}$ & \cellcolor{blue!10}\({0.126^{\pm 0.010}}\) \\ 
\hline
Backflip    &      1.69       & Failed & \cellcolor{blue!10}\({183.632^{\pm 18.536}}\) & Failed & \cellcolor{blue!10}\({0.048^{\pm 0.000}}\) & Failed & \cellcolor{blue!10}\({0.205^{\pm 0.004}}\) \\ 
\hline
Sideflip    &      2.38       & $336.549^{\pm 62.708}$ & \cellcolor{blue!10}\({166.156^{\pm 6.179}}\) & $0.051^{\pm 0.002}$ & \cellcolor{blue!10}\({0.050^{\pm 0.001}}\) & \cellcolor{blue!10}\({0.241^{\pm 0.003}}\) & $0.242^{\pm 0.005}$ \\ 
\hline
Speed Vault    &      1.92       & $388.978^{\pm 49.430}$ & \cellcolor{blue!10}\({122.465^{\pm 6.179}}\) & $0.024^{\pm 0.001}$ & \cellcolor{blue!10}\({0.023^{\pm 0.000}}\) & $0.118^{\pm 0.003}$ & \cellcolor{blue!10}\({0.116^{\pm 0.001}}\) \\ 
\hline
Double Kong    &      5.17       & Failed & \cellcolor{blue!10}\({288.489^{\pm 10.702}}\) & Failed & \cellcolor{blue!10}\({0.031^{\pm 0.001}}\) & Failed & \cellcolor{blue!10}\({0.157^{\pm 0.003}}\) \\ 
\hline
\end{tabular}
\end{table*}

\begin{table*}[!ht]
\scriptsize
\setlength{\tabcolsep}{4pt}
\renewcommand{\arraystretch}{1.5} 
\centering
\caption{Quantitative results in mean $\pm1$ std across 3 random seeds of AMP~\cite{peng2021amp} and our method in style mimicry task.}
\label{tab:AMP_singleclip_table}

\begin{tabular}{|c|c|c|c|c|c|c|c|}
\hline

\textbf{Skill} & \textbf{Length} & \multicolumn{2}{c|}{\textbf{Samples ($SR \geq 80\%$ ) [M]} } & \multicolumn{2}{c|}{\textbf{Position Error [m]}} & \multicolumn{2}{c|}{\textbf{Rotation Error [rad]}} \\ \hline

               &                 & \textbf{AMP} & \textbf{Ours} & \textbf{AMP} & \textbf{Ours} & \textbf{AMP} & \textbf{Ours} \\ 
\hline
Walk    &      1.03       &
$91.881^{\pm 0.000}$ & \cellcolor{blue!10}\({65.667^{\pm 0.000}}\) &
\cellcolor{blue!10}\({0.228^{\pm 0.002}}\) & \cellcolor{blue!10}\({0.228^{\pm 0.002}}\) &
\cellcolor{blue!10}\({0.301^{\pm 0.002}}\) & $0.304^{\pm 0.003}$ \\ 
\hline
Run    &      0.80       & 
$104.989^{\pm 0.000}$ & \cellcolor{blue!10}\({87.512^{\pm 6.179}}\) & 
$0.601^{\pm 0.004}$ & \cellcolor{blue!10}\({0.593^{\pm 0.002}}\) & 
\cellcolor{blue!10}\({0.708^{\pm 0.000}}\) & $0.718^{\pm 0.004}$\\ 
\hline
Jump    &      1.75       &
$323.442^{\pm 26.933}$ & \cellcolor{blue!10}\({196.739^{\pm 21.404}}\) &
$0.189^{\pm 0.010}$ & \cellcolor{blue!10}\({0.163^{\pm 0.003}}\) &
$0.490^{\pm 0.011}$ & \cellcolor{blue!10}\({0.480^{\pm 0.003}}\) \\ 
\hline
Spinkick    &      1.28       &
$183.632^{\pm 0.000}$ & \cellcolor{blue!10}\({161.787^{\pm 6.179}}\) &
$0.319^{\pm 0.002}$ & \cellcolor{blue!10}\({0.316^{\pm 0.008}}\) &
\cellcolor{blue!10}\({1.634^{\pm 0.003}}\) & $1.637^{\pm 0.003}$ \\ 
\hline
Cartwheel    &      2.72       &
$240.430^{\pm 6.179}$ & \cellcolor{blue!10}\({209.846^{\pm 74.914}}\) &
$0.455^{\pm 0.004}$ & \cellcolor{blue!10}\({0.399^{\pm 0.013}}\) &
\cellcolor{blue!10}\({1.604^{\pm 0.008}}\) & $1.607^{\pm 0.039}$ \\ 
\hline
Sideflip    &      2.38       &
$393.347^{\pm 28.315}$ & \cellcolor{blue!10}\({345.287^{\pm 40.517}}\) &
\cellcolor{blue!10}\({0.327^{\pm 0.010}}\) & $0.336^{\pm 0.005}$ &
\cellcolor{blue!10}\({1.254^{\pm 0.028}}\) & $1.282^{\pm 0.022}$ \\ 
\hline

\end{tabular}
\vspace{-2mm}
\end{table*}

\begin{table*}[!ht]
\scriptsize
\setlength{\tabcolsep}{4pt}
\renewcommand{\arraystretch}{1.5} 
\centering
\caption{General motion tracking performance comparison across different episode lengths. \textbf{Ours} is trained with ADD~\cite{zhang2025physics} augmented by the pose-prior reward. \textbf{Ours*} is trained in two stages: 4 billion samples guided by PDF-HR, then 8 billion samples of resumed training with the original ADD. All policies are trained with 12 billion total samples for a fair comparison.}
\label{tab:ADD_lafan_general}

\begin{tabular}{|c|c|c|c|c|c|c|c|c|c|c|c|c|}
\hline

\textbf{Ep. Length [s]} & 
\multicolumn{3}{c|}{\textbf{Joint Position Error [m]}} & 
\multicolumn{3}{c|}{\textbf{Root Position Error [m]}} & 
\multicolumn{3}{c|}{\textbf{Joint Rotation Error [rad]}} & 
\multicolumn{3}{c|}{\textbf{Root Rotation Error [rad]}} \\
\hline

   & 
\textbf{ADD} & \textbf{Ours} & \textbf{Ours*} & 
\textbf{ADD} & \textbf{Ours} & \textbf{Ours*} & 
\textbf{ADD} & \textbf{Ours} & \textbf{Ours*} & 
\textbf{ADD} & \textbf{Ours} & \textbf{Ours*}\\ 
\hline
10  & 
0.022 & 0.023 & \cellcolor{blue!10}{0.019} & 
\cellcolor{blue!10}{0.095} & 0.101 & 0.107 & 
0.121 & 0.129 & \cellcolor{blue!10}{0.108} & 
0.075 & 0.080 & \cellcolor{blue!10}{0.066} \\
\hline

20  & 
0.022 & 0.023 & \cellcolor{blue!10}{0.019} & 
\cellcolor{blue!10}{0.098} & 0.099 & 0.111 & 
0.123 & 0.127 & \cellcolor{blue!10}{0.107} & 
0.075 & 0.079 & \cellcolor{blue!10}{0.066} \\
\hline

30 &
0.022 & 0.022 & \cellcolor{blue!10}{0.018} & 
\cellcolor{blue!10}{0.103} & 0.104 & 0.112 & 
0.122 & 0.125 & \cellcolor{blue!10}{0.102} & 
0.077 & 0.077 & \cellcolor{blue!10}{0.062} \\
\hline
\end{tabular}
\vspace{-3mm}
\end{table*}
\begin{table}[!ht]
\scriptsize
\setlength{\tabcolsep}{4pt}
\renewcommand{\arraystretch}{1.5} 
\centering
\caption{Quantitative comparison of retargeting quality. We report tracking metrics from an ADD policy trained on each retargeted dataset. \textbf{MT. Length} denotes motion tracking length in frames, \textbf{PE} denotes position error, \textbf{RE} denotes rotation error, and \textbf{Deployment SR} denotes deployment success rate.}
\label{tab:ADD_amass_general}

\begin{tabular}{|c|c|c|c|c|c|c|}
\hline

\textbf{Retargeting Methods} &  
\textbf{MT. Length (frames)} &
\textbf{Body PE [m]} &
\textbf{Root PE [m]} & 
\textbf{Body RE [rad]} &
\textbf{Root RE [rad]} &
\textbf{Deployment SR}
\\
\hline

GMR & 95.708 & \cellcolor{blue!10}{0.029} & \cellcolor{blue!10}{0.113} & 0.197 & \cellcolor{blue!10}{0.087} & 6\ /\ 10 \\
\hline
Ours & \cellcolor{blue!10}{101.613} & \cellcolor{blue!10}{0.029} & 0.124 & \cellcolor{blue!10}{0.162} & \cellcolor{blue!10}{0.087} & \cellcolor{blue!10}{9\ /\ 10} \\
\hline
\end{tabular}
\vspace{-3mm}
\end{table}

\noindent
\textbf{Results.} Table~\ref{tab:ADD_singleclip_table} presents the quantitative comparisons. Our method achieves substantially higher sample efficiency across all motions and achieves lower tracking error on most hard and parkour motions. Notably, this advantage in sample efficiency becomes more pronounced as the motion difficulty increases.
The action distribution heatmap during training stage is provided in Appendix E. Compared to ADD, our method produces an action distribution that aligns much more closely with the reference motion. This matters because a humanoid’s action space is continuous and high-dimensional, while the set of actions that stay on a plausible motion manifold occupies only a small fraction of the total volume. For highly dynamic motions, the local neighborhood is dominated by invalid poses, causing ADD to waste substantial samples exploring unproductive regions. Our pose-prior guidance alleviates this by steering exploration toward a valid, well-shaped subspace. Meanwhile, since the strict tracking objective is maintained, the policy explores primarily within this reduced space, enabling faster convergence to the optimal solution. We further provide qualitative visual comparisons in Figure~\ref{fig:ADD_single_clip} to corroborate this observation.

\subsection{Style-based Motion Mimicry}
\label{subsec:style_tracking_exp}
\noindent
\textbf{Experiment settings.} We compare our method with AMP~\cite{peng2021amp} on three easy motions (Walk, Run, Jump) and three hard motions (Spinkick, Cartwheel, Sideflip). All policies are trained with 800 million samples for a fair comparison. Although style-based motion mimicry does not aim for precise motion tracking, we still report tracking error to facilitate comparison.

\noindent
\textbf{Results.} As summarized in Table~\ref{tab:AMP_singleclip_table}, our method consistently outperforms AMP in terms of sample efficiency while maintaining a comparable level of tracking error. The qualitative results at an intermediate training stage (350 million samples for all policies) are presented in Appendix E. Even with this limited training budget, our method generally mimics all reference motions, highlighting the stability and efficiency brought by our pose-prior reward. We further evaluate our method on the \textit{Target Location} and \textit{Heading} tasks and obtain consistent results. Due to space limitations, we report the task-oriented results in Appendix E.


\subsection{General Motion Tracking}
\label{subsec:general_motion_tracking_exp}
\noindent
\textbf{Experiment settings.} For this task, both our method and the baseline ADD~\cite{zhang2025physics} are trained with a single general policy to track a large-scale motion dataset, utilizing a subset of LaFAN1~\cite{harvey2020robust}. The subset contains 34 sequences (excluding six FallAndGetUp sequences), totaling 7887.6 seconds of motion. We adopt a two-stage training strategy (dubbed as \textbf{Ours*}) to improve the tracking accuracy. For this strategy, we first train the policy with the pose-prior reward for 4 billion samples, then disable the prior and resume the training with the original ADD objective. All policies are trained with 12 billion samples in total for a fair comparison. To assess convergence under different horizon lengths, we additionally vary the episode duration and evaluate tracking horizons of 10, 20, and 30 seconds for both our method and ADD.


\noindent
\textbf{Results.} Quantitative results are reported in Table~\ref{tab:ADD_lafan_general}, and learning curves are provided in Appendix E. Benefiting from the guidance of the PDF-HR pose-prior reward, our method exhibits significantly faster convergence compared to the baseline. This advantage becomes more pronounced as the episode length increases, demonstrating improved robustness for long-horizon tracking. We also observe that the one-stage training yields slightly higher tracking error than ADD, whereas the two-stage protocol effectively mitigates this issue and consistently achieves the lowest joint position error, joint rotation error, and root rotation error. Intuitively, the first stage uses the prior reward to quickly acquire basic locomotion and coordination skills, as reflected by the rapid rise in success rate; the second stage then focuses purely on tracking to refine accuracy. Overall, the two-stage method strikes a strong balance between fast convergence and precise motion tracking.

\subsection{Motion Retargeting}
\label{subsec:motion_retargeting_exp}
\noindent
\textbf{Experiment settings.}
For the motion retargeting task, we compare our method with GMR~\cite{araujo2025retargeting} on a subset of the AMASS~\cite{AMASS:ICCV:2019} DanceDB dataset. 30 sequences are sampled from DanceDB, totaling 7,770 frames (254.7 seconds). We retarget the same SMPL-X~\cite{SMPL-X:2019} motions to the Unitree G1 robot using both GMR and our method, producing two sets of retargeted reference motions. To evaluate how well these motions can be tracked in physics, we train an ADD-based general tracking policy on each reference set and compare the resulting tracking errors.

\begin{figure}[t]
\centering
\begin{minipage}{0.54\textwidth}
\centering
\includegraphics[width=\linewidth]{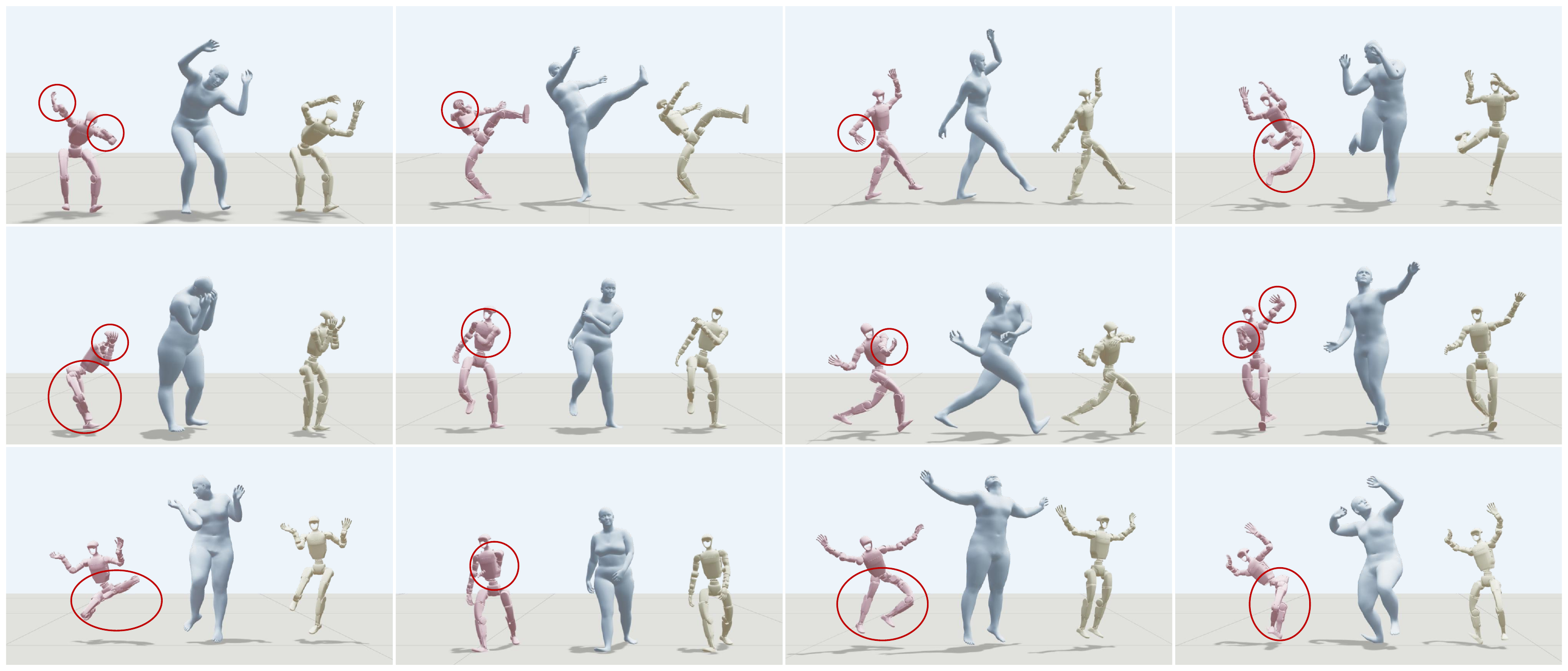}
\caption{
Visual comparison of retargeted motions produced by GMR~\cite{araujo2025retargeting} (red) and our method (green). GMR artifacts are highlighted with red markers.
}
\label{fig:amass_retargeting_cmp} 
\end{minipage}
\hfill
\begin{minipage}{0.45\textwidth}
\centering
\includegraphics[width=\linewidth]{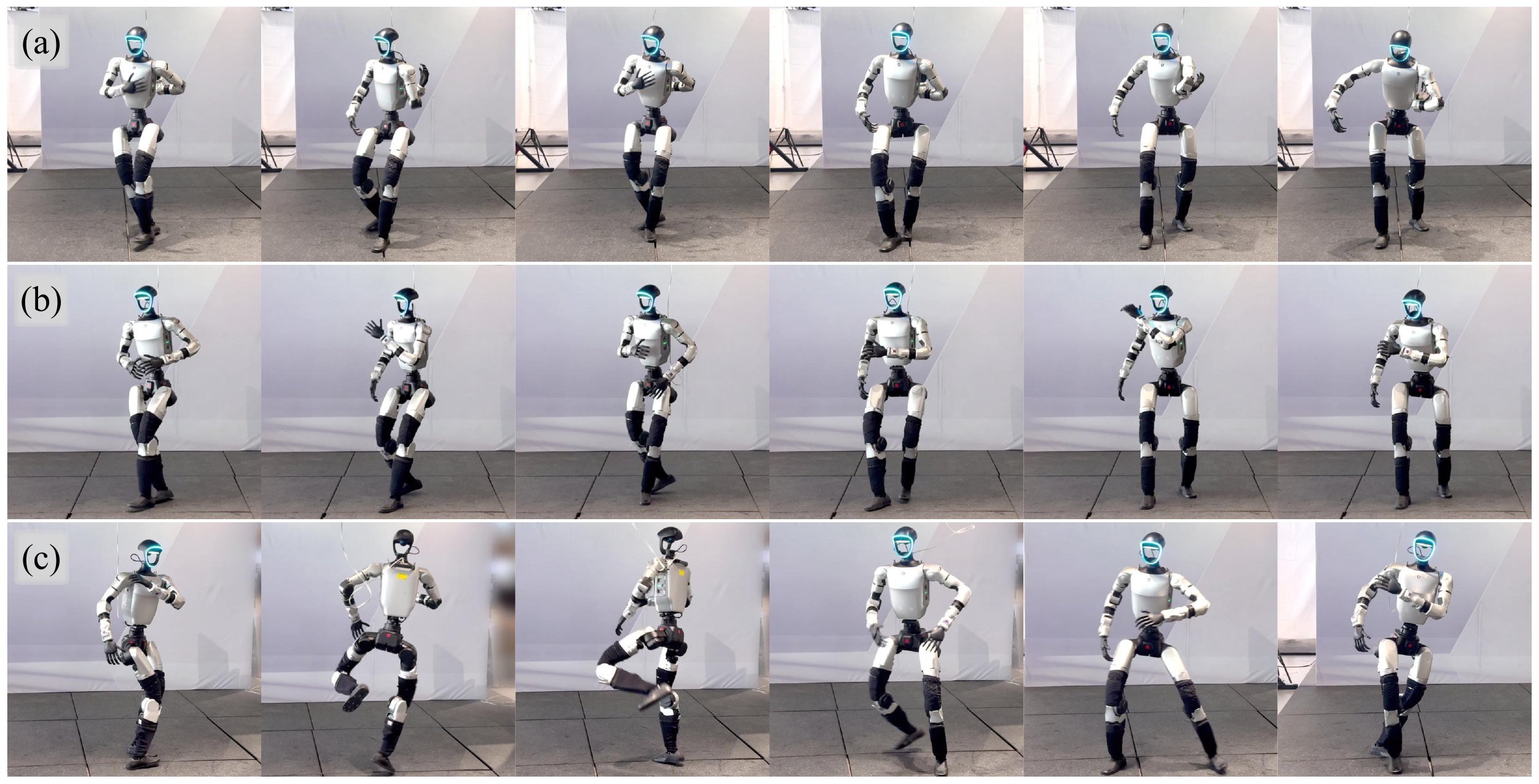}
\caption{
\textbf{Real-world deployment.} (a) GMR on Seq. 4. (b) Ours on Seq. 4. (c) presents our results on Seq. 2 for highly dynamic skills, where GMR fails.
}
\label{fig:beyondmimic_real}
\end{minipage}
\vspace{-3mm}
\end{figure}

\noindent
\textbf{Results.} Figure~\ref{fig:amass_retargeting_cmp} presents a visual comparison between the retargeted motions and the original SMPL-X motions. As highlighted by the red indicators, GMR often produces kinematically implausible artifacts, including severe joint distortions (e.g., at the wrists, torso, and legs) and noticeable self-penetration. These artifacts not only deviate from the original human motion but are also physically infeasible, which makes them difficult to track in physics-based simulators. The quantitative results in Table~\ref{tab:ADD_amass_general} corroborate this observation: such inconsistencies in GMR lead to shorter continuous tracking durations and larger body rotation errors. In contrast, our PDF-HR regularizer steers the optimizer away from poor local minima without requiring tedious parameter tuning, yielding retargeted motions that are both visually faithful to the reference and physically valid for tracking.

\noindent

\begin{figure}[!t] 
\centering
\includegraphics[width=0.99\linewidth]{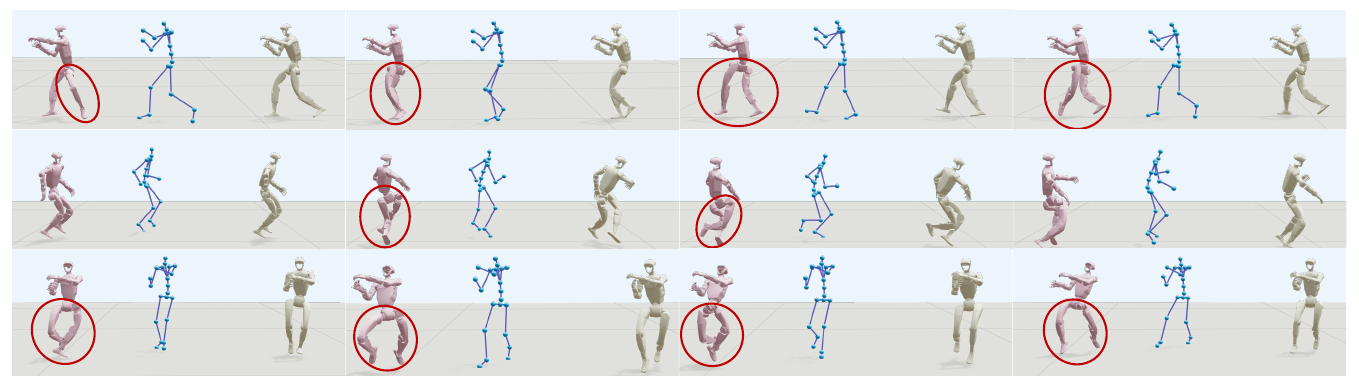}
\caption{Visual comparison of retargeted motions produced by OmniRetarget~\cite{yang2025omniretarget} (red) and our method (green). OmniRetarget artifacts are high-lighted with red markers.}
\label{fig:omni_sup_vis} 
\end{figure}


\textbf{Deployment validation.}
We further evaluate deployability with BeyondMimic~\cite{liao2025beyondmimic} on 10 motion sequences, comparing policies trained on GMR-retargeted data and our retargeted data. Our method successfully deploys on 9/10 sequences, compared with 6/10 for GMR. As shown in Figure~\ref{fig:beyondmimic_real}, our retargeted motions yield more natural behaviors that better match the targets. Figure~\ref{fig:beyondmimic_real} (c) further highlights highly dynamic skills, such as dance motions and spinning jumps, where our method remains robust while GMR often fails. Detailed learning curves are provided in Appendix E.


To demonstrate the generalizability of our method, we further integrate PDF-HR into OmniRetarget~\cite{yang2025omniretarget}. We provide quantitative and qualitative comparisons in Figure~\ref{fig:omni_sup_vis} and Table~\ref{tab:omni_quantitive}. All results are obtained using the official implementation of OmniRetarget on the LaFAN1~\cite{harvey2020robust} dataset, following the official evaluation protocol. Our method achieves consistently better results.
\begin{table}[!ht]
\scriptsize
\setlength{\tabcolsep}{4pt}
\renewcommand{\arraystretch}{1.5} 
\centering
\caption{Comparison of physical constraints and optimization costs. \textbf{Baseline (OmniRetarget)} represents the SOTA approach, while \textbf{Ours} incorporates the proposed method. Lower values indicate better physical realism.}
\label{tab:omni_quantitive}

{
\begin{tabular}{|l|c|c|}
\hline
\textbf{Metric} & \textbf{Baseline (OmniRetarget)} & \textbf{OmniRetarget + Ours} \\
\hline
Penetration Duration [Ratio] & $0.0009 \pm 0.0031$ & \cellcolor{blue!10}{$0.0005 \pm 0.0017$} \\
\hline
Max Penetration [m] & $0.0124 \pm 0.0015$ & \cellcolor{blue!10}{$0.0122 \pm 0.0014$} \\
\hline
Sliding Duration [Ratio] & $0.0000 \pm 0.0000$ & $0.0000 \pm 0.0000$ \\
\hline
Max Sliding Velocity [m/s] & $0.0000 \pm 0.0000$ & $0.0000 \pm 0.0000$ \\
\hline
Optimization Cost [-] & $0.1311 \pm 0.1122$ & \cellcolor{blue!10}{$0.1147 \pm 0.1391$} \\
\hline
\end{tabular}
}
\end{table}

\section{Conclusion} 
\label{sec:conclusion}
In this work, we introduce Pose Distance Fields for Humanoid Robots (PDF-HR), a lightweight and differentiable prior that models the plausibility of humanoid robot poses as a continuous distance field learned from a large corpus of retargeted motions. PDF-HR is simple to integrate into existing pipelines and broadly applicable to optimization and learning. Across multiple humanoid benchmarks, including single-trajectory tracking, general motion tracking, style-conditioned mimicry, and motion retargeting, PDF-HR consistently improves strong baselines, demonstrating the value of pose-level priors for robust humanoid behavior. Our results highlight the importance of pose priors for humanoid control and open opportunities for more data-driven approaches in humanoid motion generation and understanding.

\section{Limitations}
\label{sec:limitations}
While PDF-HR demonstrates promising results across a range of humanoid tasks, it has several limitations:
\begin{itemize}[leftmargin=*]
    \item \textbf{Accuracy in some tracking settings.} Although PDF-HR can substantially accelerate convergence, its final tracking accuracy is occasionally lower than that of the baseline. We acknowledge this as a limitation.
    \item \textbf{Mode collapse in style-conditioned tracking.} In style tracking, our method can suffer from mode collapse, similar to AMP~\cite{peng2021amp}. A potential remedy is to enrich the prior with additional kinematic information, e.g., velocity- and acceleration-based fields~\cite{yu2025geometric}, to better constrain long-horizon behavior.
    \item \textbf{Runtime and smoothness in retargeting.} Our current implementation is slower than GMR~\cite{araujo2025retargeting}, and it can sometimes exhibit more severe jitter. GPU-parallel batch processing and incorporating explicit smoothness costs could mitigate these issues and better reveal the practical advantages of our approach~\cite{kim2025pyroki}.
    \item \textbf{Training data quality.} Our pose field is trained on retargeted poses, which may not always be physically feasible. An important future direction is to curate or collect large-scale pose corpora that better satisfy physical constraints, thereby improving the fidelity and robustness of the learned field.
\end{itemize}

\begin{ack}
We gratefully acknowledge the financial support provided by BrainCo.
This work is partially supported by the Research Travel Grant of the
Base of Red Bird MPhil (RBM) at The Hong Kong University of Science
and Technology (Guangzhou).
We would like to thank our RBM for the academic support.
\end{ack}


\bibliographystyle{plainnat}
\bibliography{references}

\appendix
\newpage
\appendix\label{appendix}

\section{PDF-HR Training Details}
\label{appendix:pdf_training_details}

\noindent To learn a continuous and differentiable representation of the pose distance field, dense coverage of the state space is essential. Pose-NDF~\cite{tiwari2022pose} typically generates off-manifold samples by perturbing clean poses with Gaussian noise. Formally, given a clean pose $\mathbf{q}$ on the manifold, a corrupted sample is generated as $\hat{\mathbf{q}} = \mathbf{q} + \boldsymbol{\epsilon}$, where $\boldsymbol{\epsilon} \sim \mathcal{N}(0, \sigma^2\mathbf{I})$. However, this method introduces a sampling bias. In an $N$-dimensional space, the normalized squared Euclidean norm of the noise vector follows a Chi-squared distribution:\begin{equation}\frac{|\boldsymbol{\epsilon}|^2}{\sigma^2} \sim \chi^2_N.\end{equation}

Consequently, the perturbation magnitudes concentrate within a narrow spherical shell around $\sigma\sqrt{N}$, preventing the network from effectively learning the distance field's transition across different scales. This concentration compromises the continuity of the represented manifold. To address this limitation, we construct a composite training dataset $\mathcal{D}$ derived from a high-fidelity reference dataset, denoted as $\mathcal{D}_{\text{raw}}$. The final training set $\mathcal{D} = \mathcal{D}_{\text{on}} \cup \mathcal{D}_{\text{near}} \cup \mathcal{D}_{\text{interp}}$ aggregates samples from three distinct generation strategies.

\vspace{5pt}
\noindent \textbf{On-Manifold Sampling.} We directly sample raw pose states $\mathbf{q}$ from the reference dataset $\mathcal{D}_{\text{raw}}$. These samples represent the ground truth geometry of the valid pose distance field. Consequently, for every sample $\mathbf{q} \in \mathcal{D}_{\text{raw}}$, we assign a zero-distance label:
\begin{equation}
    \mathcal{D}_{\text{on}} = \{ (\mathbf{q}, 0) \mid \mathbf{q} \in \mathcal{D}_{\text{raw}} \}.
\end{equation}

\noindent \textbf{Near-Manifold Sampling.} To characterize the distance field gradients in the immediate vicinity of the manifold, we employ a decoupled perturbation strategy. Unlike standard Gaussian noise injection, which entangles magnitude and direction, our approach explicitly separates the perturbation direction from its scalar magnitude. This facilitates precise control over the sampling density near the manifold boundary.
Given a state $\mathbf{q} \in \mathcal{D}_{\text{raw}}$, we generate a query point $\tilde{\mathbf{q}}$ via:
\begin{equation}
\begin{aligned}
    \tilde{\mathbf{q}} &= \mathrm{clip}\left(\mathbf{q} + r \cdot \bm{u},\ \mathbf{q}_{\min},\ \mathbf{q}_{\max}\right), \\
    \text{with} \quad \bm{u} &\sim \mathrm{Unif}(\mathbb{S}^{N-1}), \quad r \sim |\mathcal{N}(0, \sigma^2)|.
\end{aligned}
\end{equation}
Here, the perturbation direction $\bm{u}$ is drawn uniformly from the unit hypersphere $\mathbb{S}^{N-1}$, and the scalar magnitude $r$ is sampled from a half-normal distribution to concentrate sampling density near the manifold. The $\mathrm{clip}$ operation ensures $\tilde{\mathbf{q}}$ remains within joint limits $[\mathbf{q}_{\min}, \mathbf{q}_{\max}]$. The supervision label for $\tilde{\mathbf{q}}$ is computed as the distance to its nearest neighbor in the training set, retrieved via FAISS:
\begin{equation}
    \mathcal{D}_{\text{near}} = \{ (\tilde{\mathbf{q}}, d) \mid d = \min_{\mathbf{q} \in \mathcal{D}_{\text{raw}}} \|\tilde{\mathbf{q}} - \mathbf{q}\| \}.
\end{equation}

\noindent \textbf{Interpolation-based Sampling.}
To ensure the learned distance field is globally smooth and continuous, we employ linear interpolation. Since the raw data consists of discrete points, we generate intermediate samples between random pairs of poses to bridge gaps in the state space. We randomly sample pairs $\mathbf{q}_a, \mathbf{q}_b$ from $\mathcal{D}_{\text{raw}}$ and construct intermediate states $\tilde{\mathbf{q}} = \alpha \mathbf{q}_a + (1-\alpha)\mathbf{q}_b$ using a mixing coefficient $\alpha \sim \mathrm{Unif}(0, 1)$. The interpolation dataset is defined as:
\begin{equation}
    \mathcal{D}_{\text{interp}} = \{ (\tilde{\mathbf{q}}, d) \mid d = \min_{\mathbf{q} \in \mathcal{D}_{\text{raw}}} \|\tilde{\mathbf{q}} - \mathbf{q}\| \}.
\end{equation}

The final dataset $\mathcal{D}$ comprises 300 million state-distance pairs $(\mathbf{q}, d)$. The data generation process was executed on a server equipped with 10 NVIDIA RTX 3090 GPUs and was completed in approximately 10 hours.

\section{Training Network}
\noindent
To explicitly embed the kinematic structure of the 29-DoF G1 humanoid into our representation, we design a network architecture that mirrors the robot's topology. Let $\mathbf{q} = [q_1, \dots, q_K]^\top$ denote the input configuration, where $q_k \in \mathbb{R}$ represents the scalar state of the $k$-th joint. We define a parent mapping $\tau: \{2, \dots, K\} \to \{1, \dots, K\}$ that returns the index of the parent joint for any given joint $k$. To capture the spatial dependency of each link on its ancestors, the network encodes features hierarchically along the kinematic chain. Formally, we employ a set of local encoders $\{f_{\text{enc}}^{(k)}\}_{k=1}^K$ to propagate latent information from the root to the end-effectors:
\begin{equation}
\begin{aligned}
    \mathbf{v}_1 &= f_{\text{enc}}^{(1)}(q_1), \\
    \mathbf{v}_k &= f_{\text{enc}}^{(k)}(q_k, \mathbf{v}_{\tau(k)}), \quad \text{for } k \in \{2, \dots, K\},
\end{aligned}
\end{equation}
where $\mathbf{v}_k \in \mathbb{R}^{16}$ denotes the latent feature for joint $k$. In this formulation, the root feature is derived purely from its local state, while subsequent child encodings are conditioned on both their local joint state $q_k$ and the parent feature $\mathbf{v}_{\tau(k)}$. 

PDF-HR is lightweight, with only 0.42M parameters. It runs at 635.5 FPS on an RTX 3090 with batch size 1, making it well suited for RL training with frequent simulator interactions.

\section{Riemannian Gradient for 1-DoF Joints}
\label{appendix:RG_proof}
\noindent
\textbf{Definition 1} (Tangent Space)\textbf{.} 
To characterize motion on the curved manifold $\mathcal{M}$ embedded in $\mathcal{X}$, we introduce the tangent space $T_{\mathbf{x}}\mathcal{M}$. Formally, a vector $\mathbf{v} \in \mathcal{X}$ is tangent to $\mathcal{M}$ at $\mathbf{x}$ if it is the velocity of a smooth curve passing through $\mathbf{x}$. Let $\gamma: [0, 1] \to \mathcal{M}$ be a smooth curve such that $\gamma(0) = \mathbf{x}$. The tangent space is defined as the collection of all such velocities:

\begin{equation}\label{eq:tangent_space}
    T_{\mathbf{x}}\mathcal{M} \triangleq \{ \dot{\gamma}(0) \mid \gamma \text{ is smooth}, \gamma(0) = \mathbf{x} \} \subset \mathcal{X}.
\end{equation}
where $\dot{\gamma}(0)$ is understood as an element of $T_{\iota(\mathbf{x})} \mathcal{X}$ via the differential of the embedding $\iota: \mathcal{M} \hookrightarrow \mathcal{X}$. While the ambient space $\mathcal{X}$ supports global linear operations, rigorous kinematic updates must respect the manifold structure defined on the tangent bundle, the disjoint union of tangent spaces: $T\mathcal{M} \triangleq \bigcup_{\mathbf{x} \in \mathcal{M}} \{ (\mathbf{x}, \mathbf{v}) \mid \mathbf{v} \in T_{\mathbf{x}}\mathcal{M} \}$.

\noindent
\textbf{Definition 2} (Exponential Map on $\mathrm{SO}(3)$)\textbf{.}
The matrix exponential $\exp:\mathfrak{so}(3)\rightarrow \mathrm{SO}(3)$ maps a Lie algebra element to a rotation.
Given a unit axis $\hat{\boldsymbol{\omega}}\in\mathbb{R}^3$ and angle $\theta\in\mathbb{R}$, Rodrigues' formula yields
\begin{equation}\label{eq:exp_map}
\exp([\hat{\boldsymbol{\omega}}]\theta)
=
\mathbf{I}
+\sin\theta\,[\hat{\boldsymbol{\omega}}]
+(1-\cos\theta)\,[\hat{\boldsymbol{\omega}}]^2,
\end{equation}
where $[\cdot]$ denotes the skew-symmetric matrix operator.

\noindent
\textbf{Definition 3} (Local Logarithmic Map on $\mathrm{SO}(3)$)\textbf{.}
The matrix logarithm $\log:\mathrm{SO}(3)\rightarrow \mathfrak{so}(3)$ is the local inverse of $\exp$.
For two rotations $\mathbf{R}_1,\mathbf{R}_2\in\mathrm{SO}(3)$, define $\mathbf{R}\triangleq \mathbf{R}_1^\top\mathbf{R}_2$.
Let $\theta\in[0,\pi]$ satisfy $\cos\theta=\frac{\mathrm{tr}(\mathbf{R})-1}{2}$.
Then (for $\theta\notin\{0,\pi\}$), the principal logarithm is given by
\begin{equation}
[\boldsymbol{\omega}]
=
\log(\mathbf{R})
=
\frac{\theta}{2\sin\theta}\left(\mathbf{R}-\mathbf{R}^\top\right).
\end{equation}
The geodesic distance induced by the canonical bi-invariant Riemannian metric on $\mathrm{SO}(3)$ is given by
\begin{equation}\label{eq:geo_def_sup}
d_{\mathrm{geo}}(\mathbf{R}_1,\mathbf{R}_2)
=
\left\|\big(\log(\mathbf{R}_1^\top\mathbf{R}_2)\big)^{\vee}\right\|_2,
\end{equation}
where $(\cdot)^{\vee}:\mathfrak{so}(3)\rightarrow\mathbb{R}^3$ is the inverse of the skew operator, and $\|\cdot\|_2$ is the Euclidean norm.

Let $f:\mathrm{SO}(3)\!\rightarrow\!\mathbb{R}$ be differentiable, and let $\nabla f(\mathbf{R})$ denote the Euclidean (ambient) gradient under the Frobenius inner product $\langle \mathbf{A},\mathbf{B}\rangle \triangleq \mathrm{Tr}(\mathbf{A}^\top \mathbf{B})$. Projecting $\nabla f(\mathbf{R})$ onto the tangent space $T_{\mathbf{R}}\mathrm{SO}(3)$ yields the Riemannian gradient:
\begin{align}
\mathrm{grad}\, f(\mathbf{R})
&\triangleq \Pi_{\mathbf{R}}\!\left(\nabla f(\mathbf{R})\right) \nonumber\\
&= \mathbf{R}\,\mathrm{skew}\!\left(\mathbf{R}^\top \nabla f(\mathbf{R})\right) \nonumber\\
&= \frac{1}{2}\mathbf{R}\!\left(\mathbf{R}^\top \nabla f(\mathbf{R}) - \nabla f(\mathbf{R})^\top \mathbf{R}\right) \nonumber\\
&= \frac{1}{2}\nabla f(\mathbf{R}) - \frac{1}{2}\mathbf{R}\nabla f(\mathbf{R})^\top \mathbf{R},
\label{eq:riem_grad_so3}
\end{align}
where $\Pi_{\mathbf{R}}:\mathbb{R}^{3\times 3}\!\rightarrow\!T_{\mathbf{R}}\mathrm{SO}(3)$ is the orthogonal projector and
$\mathrm{skew}(\mathbf{A})\triangleq(\mathbf{A}-\mathbf{A}^\top)/2$.

\noindent
\textbf{Restriction to a 1-DoF revolute joint.}
For a 1-DoF joint with fixed unit axis $\mathbf{k}\in\mathbb{R}^3$, the admissible motions form a 1D submanifold whose tangent space is spanned by
\begin{equation}
\mathbf{V} \triangleq \mathbf{R}[\mathbf{k}] \in T_{\mathbf{R}}\mathrm{SO}(3).
\end{equation}
To compare the gradient signal along this 1D direction, we compute the inner product between the second term of~\eqref{eq:riem_grad_so3} and $\mathbf{V}$:
\begin{align}
&\left\langle -\frac{1}{2}\mathbf{R}\nabla f(\mathbf{R})^\top \mathbf{R},\, \mathbf{R}[\mathbf{k}] \right\rangle\\
&= \mathrm{Tr}\!\left(\left(-\frac{1}{2}\mathbf{R}\nabla f(\mathbf{R})^\top \mathbf{R}\right)^\top \mathbf{R}[\mathbf{k}]\right) \nonumber\\
&= -\frac{1}{2}\mathrm{Tr}\!\left(\mathbf{R}^\top \nabla f(\mathbf{R})\, \mathbf{R}^\top \mathbf{R}[\mathbf{k}]\right) \nonumber\\
&= -\frac{1}{2}\mathrm{Tr}\!\left(\mathbf{R}^\top \nabla f(\mathbf{R})[\mathbf{k}]\right).
\label{eq:term2}
\end{align}
Using $[\mathbf{k}]^\top=-[\mathbf{k}]$ and the cyclic property of the trace, we can rewrite~\eqref{eq:term2} as
\begin{align}
-\frac{1}{2}\mathrm{Tr}\!\left(\mathbf{R}^\top \nabla f(\mathbf{R})[\mathbf{k}]\right)
&= \frac{1}{2}\mathrm{Tr}\!\left(\mathbf{R}^\top \nabla f(\mathbf{R})[\mathbf{k}]^\top\right) \nonumber\\
&= \frac{1}{2}\mathrm{Tr}\!\left(\left(\nabla f(\mathbf{R})^\top \mathbf{R}\right)[\mathbf{k}]\right) \nonumber\\
&= \frac{1}{2}\mathrm{Tr}\!\left(\nabla f(\mathbf{R})^\top \mathbf{R}[\mathbf{k}]\right) \nonumber\\
&= \left\langle \frac{1}{2}\nabla f(\mathbf{R}),\, \mathbf{R}[\mathbf{k}] \right\rangle.
\label{eq:term_equiv}
\end{align}
Combining~\eqref{eq:riem_grad_so3} and~\eqref{eq:term_equiv} gives
\begin{equation}
\left\langle \mathrm{grad}\, f(\mathbf{R}),\, \mathbf{R}[\mathbf{k}] \right\rangle
=
\left\langle \nabla f(\mathbf{R}),\, \mathbf{R}[\mathbf{k}] \right\rangle.
\end{equation}
Thus, for a 1-DoF revolute joint, the gradient component along the admissible tangent direction $\mathbf{R}[\mathbf{k}]$ is identical whether computed from the Riemannian gradient or directly from the Euclidean gradient. Equivalently, the derivative with respect to the scalar joint angle $q$ can be obtained using the Euclidean gradient $\nabla f(\mathbf{R})$.

\section{Implementation Details of RL-based Tracking}
\label{appendix:implementation}
All RL-based experiments, including our method, AMP~\cite{peng2021amp} and ADD~\cite{zhang2025physics}, are implemented using MimicKit~\cite{peng2025mimickit} framework. Single-trajectory and style-based tracking are run on NVIDIA RTX 3090 GPUs, whereas the large-scale general motion tracking experiments are run on NVIDIA H20 GPUs. For fairness, we use the same default MimicKit hyperparameters for all methods (ours and baselines). For experiments that include our pose-prior reward, the reward weight $w^{P}$ and scale $\alpha^{P}$ are set to $0.2$ and $1$, respectively, except for the Sideflip motion in style mimicry, where we set $w^{P}=0.05$.

The retargeting experiments are based on the GMR framework~\cite{araujo2025retargeting} using the same configurations. We implement the prior task within the Mink IK solver~\cite{Zakka_Mink_Python_inverse_2025} and solve it jointly with the GMR-defined tasks.

\section{More Experimental Results}
\label{appendix:extra_results}
\begin{figure}[!t] 
\centering
\includegraphics[width=0.7\linewidth]{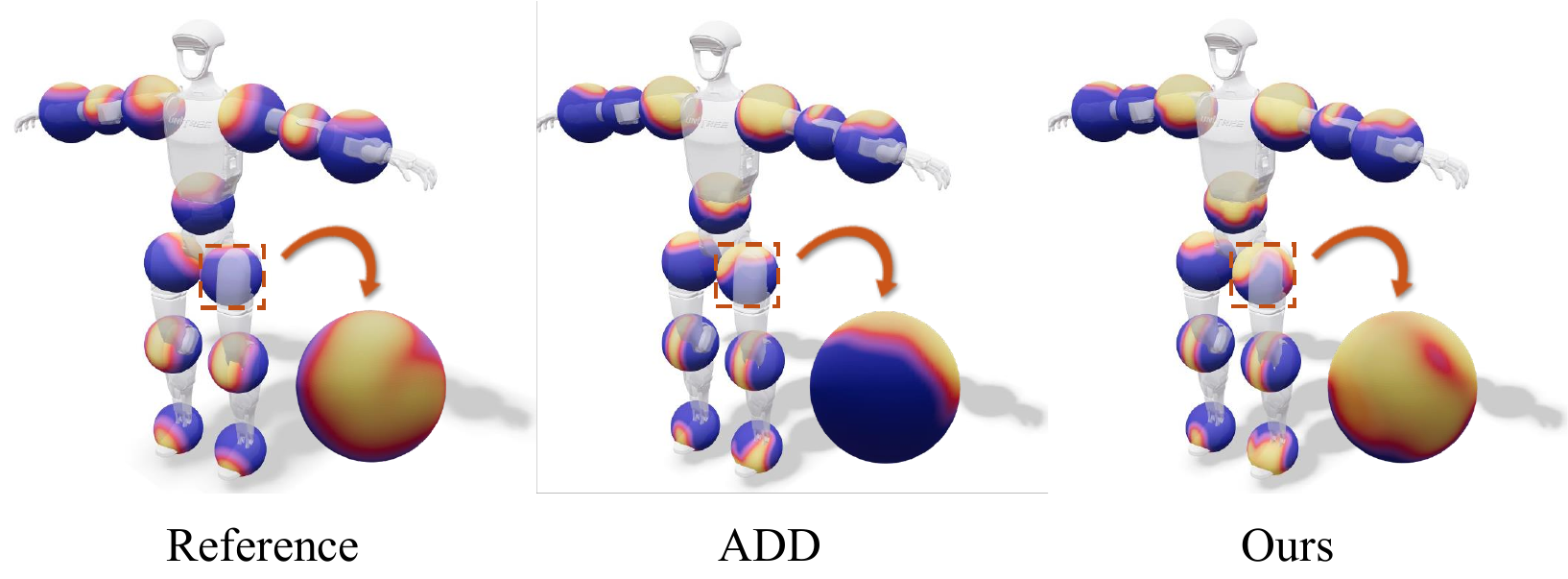}
\vspace{-2mm}
\caption{\textbf{Visualization of joint orientation distributions of Sideflip at early training stage.} The visualization maps the directional vectors of the robot's links onto unit spheres centered at their respective joints. The color gradient corresponds to the probability density of the visited states.}
\label{fig:joint_map} 
\end{figure}

\begin{figure}[!ht]
\vspace{-3mm}
\centering
\includegraphics[width=0.93\linewidth]{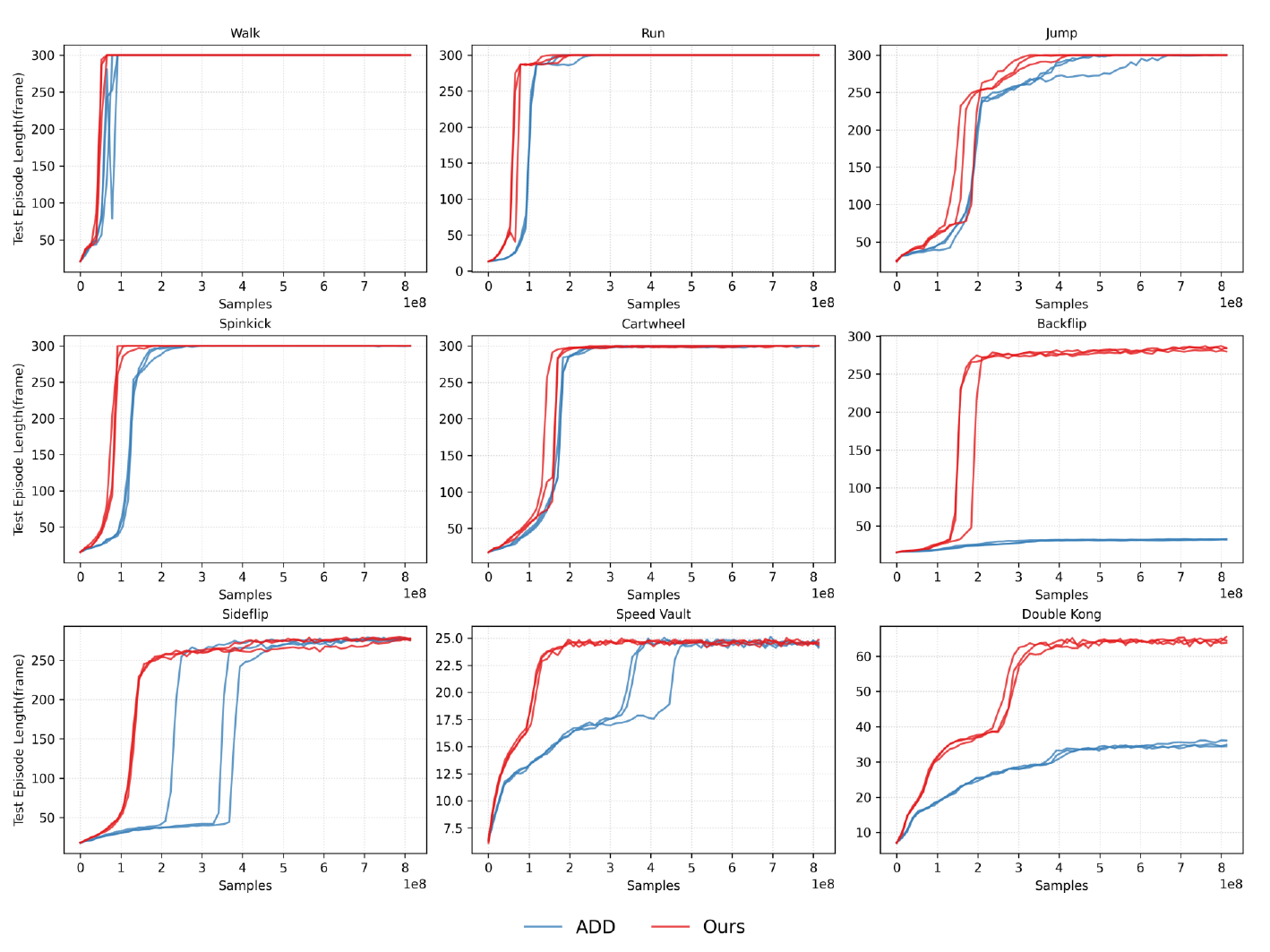}
\vspace{-3mm}
\caption{
The learning curves of ADD and our method on the single-trajectory motion tracking task.
}
\label{fig:ADD_singleclip_curve} 
\end{figure}

\begin{figure}[!ht] 
\centering
\vspace{-3mm}
\includegraphics[width=0.93\linewidth]{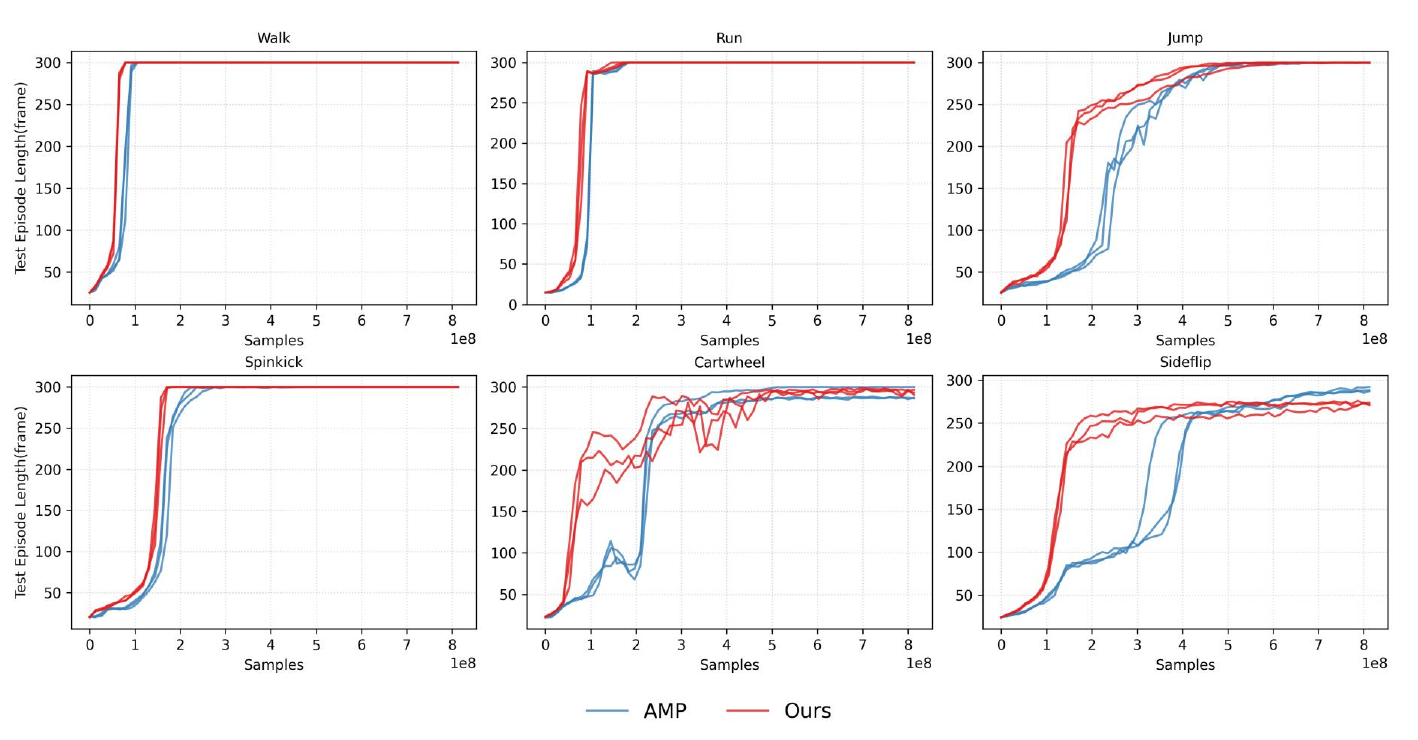}
\vspace{-3mm}
\caption{
The learning curves of AMP and our method on the style-based motion mimicry task.
}
\label{fig:AMP_singleclip_curve}

\end{figure}

\begin{figure}[!ht] 
\centering
\includegraphics[width=0.99\linewidth]{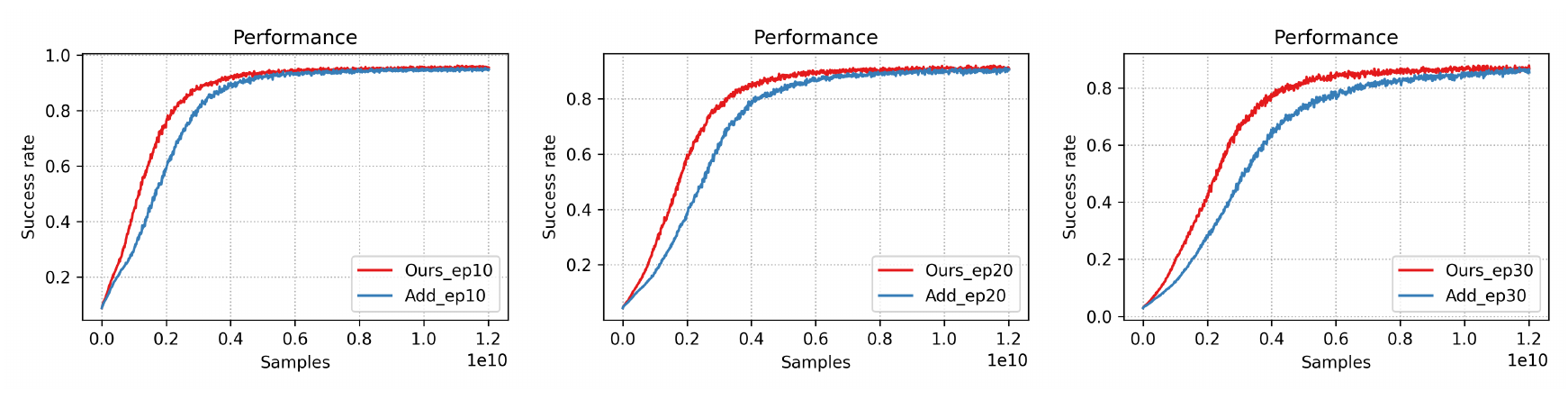}
\vspace{-3mm}
\caption{
The learning curves of ADD and our method on general motion tracking task across various episode length.
}
\label{fig:general_12} 
\end{figure}
\begin{figure}[!ht] 
\centering
\includegraphics[width=0.99\linewidth]{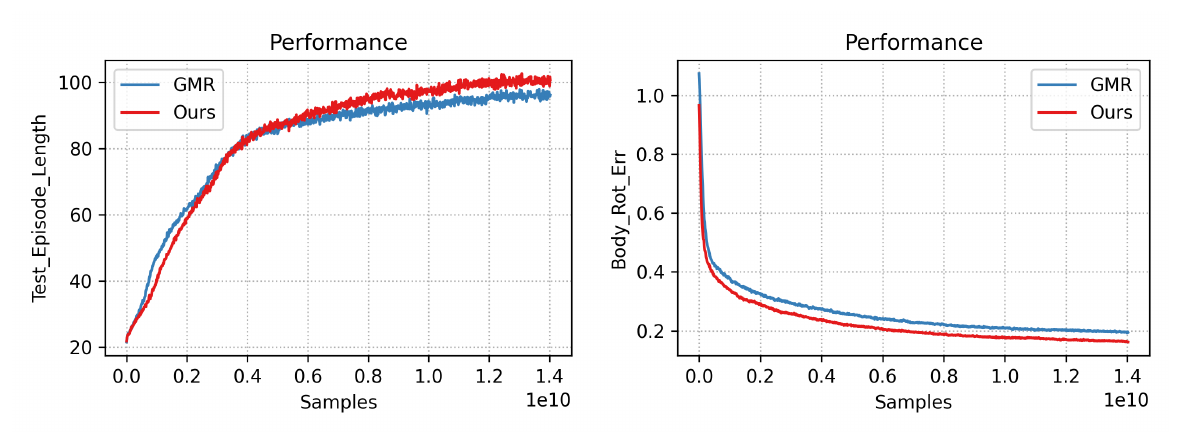}
\vspace{-3mm}
\caption{
The learning curves of the ADD general motion tracker trained on datasets produced by GMR and our method.
}
\label{fig:amass_RL_curve} 
\end{figure}

\begin{figure*}[!ht] 
\centering
\includegraphics[width=0.99\linewidth]{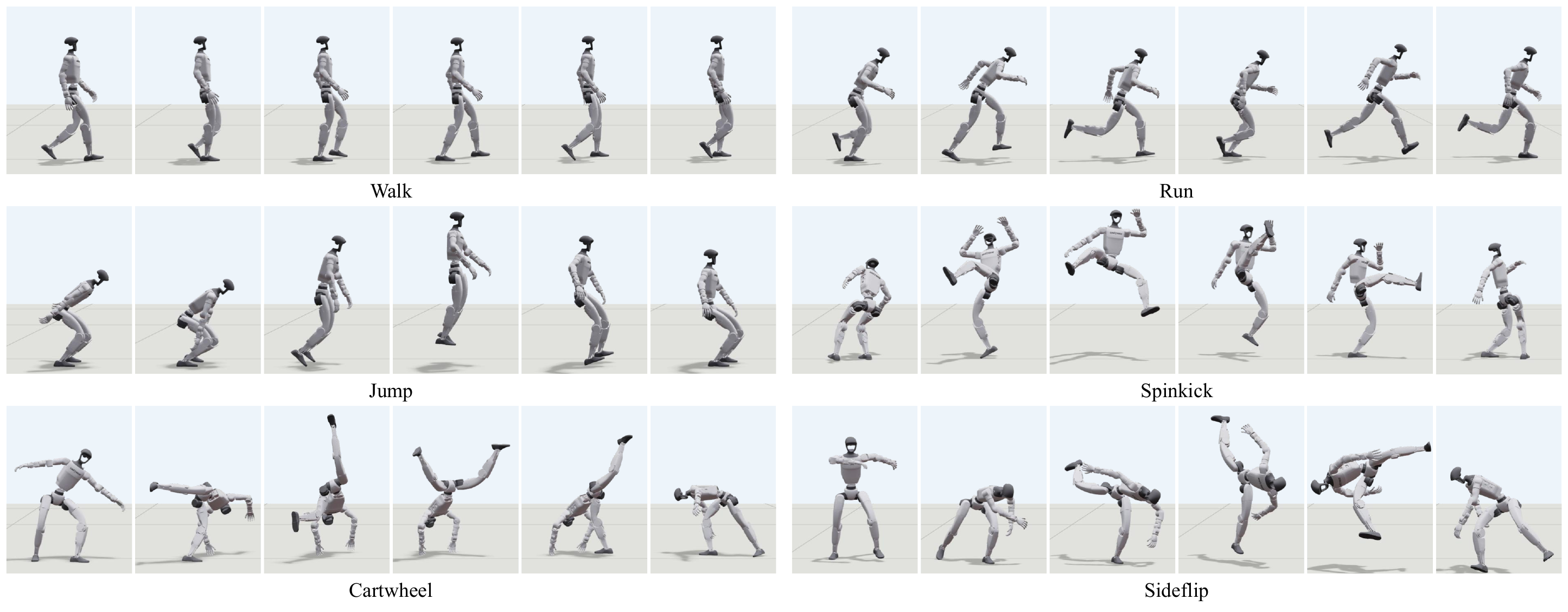}
\caption{
Qualitative results of our method in style-based motion mimicry task with only $350M$ samples.
}
\label{fig:AMP_single_clip} 
\end{figure*}

\begin{figure}[!ht] 
\centering
\includegraphics[width=0.99\linewidth]{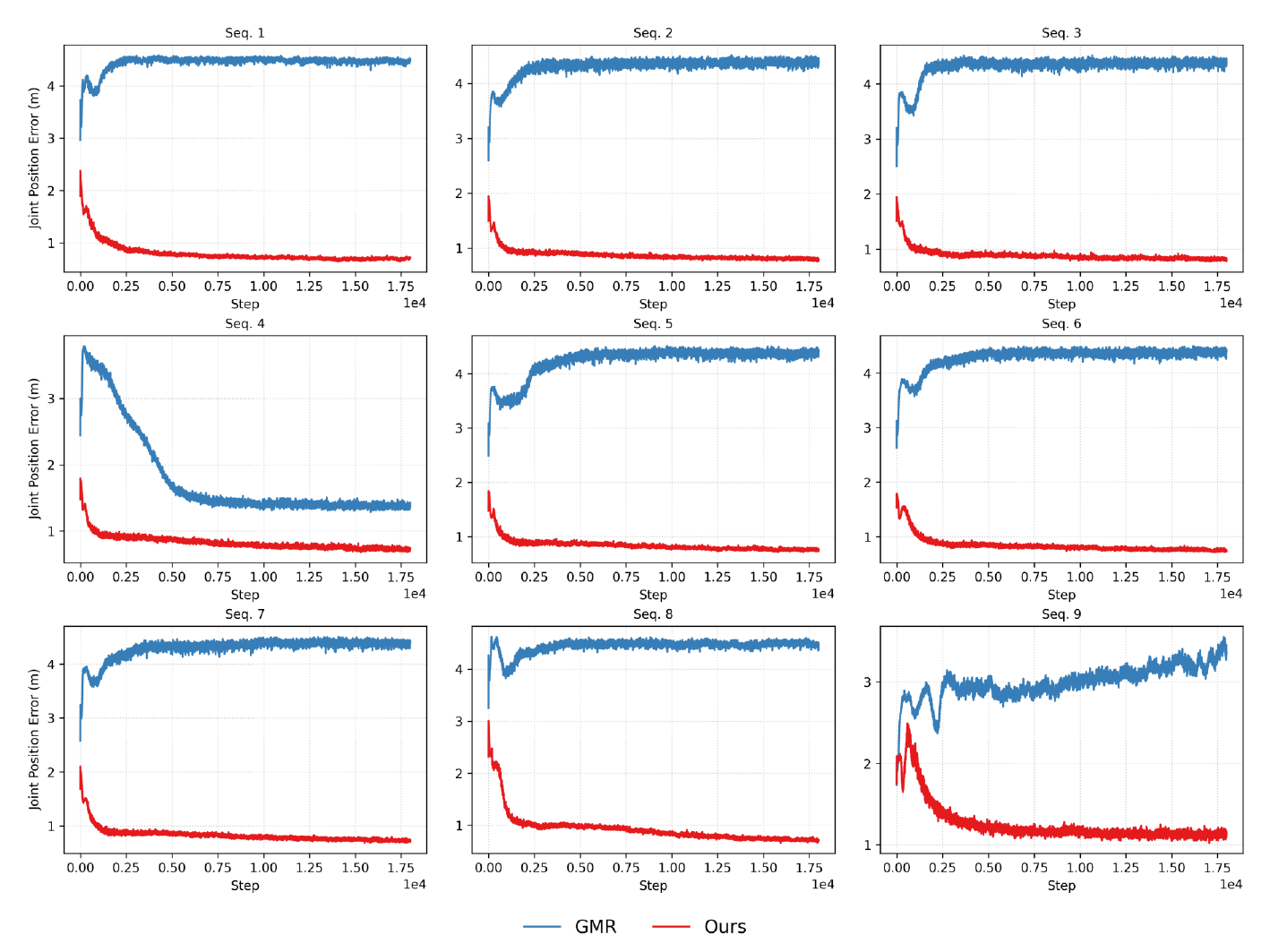}
\vspace{-4mm}
\caption{
The learning curves of joint position error for the Beyondmimic policy trained on GMR-retargeted and our retargeted motion.
}
\label{fig:beyondmimic_curve_joint_pos_error}
\vspace{-4mm}
\end{figure}

\begin{figure}[!ht] 
\centering
\includegraphics[width=0.99\linewidth]{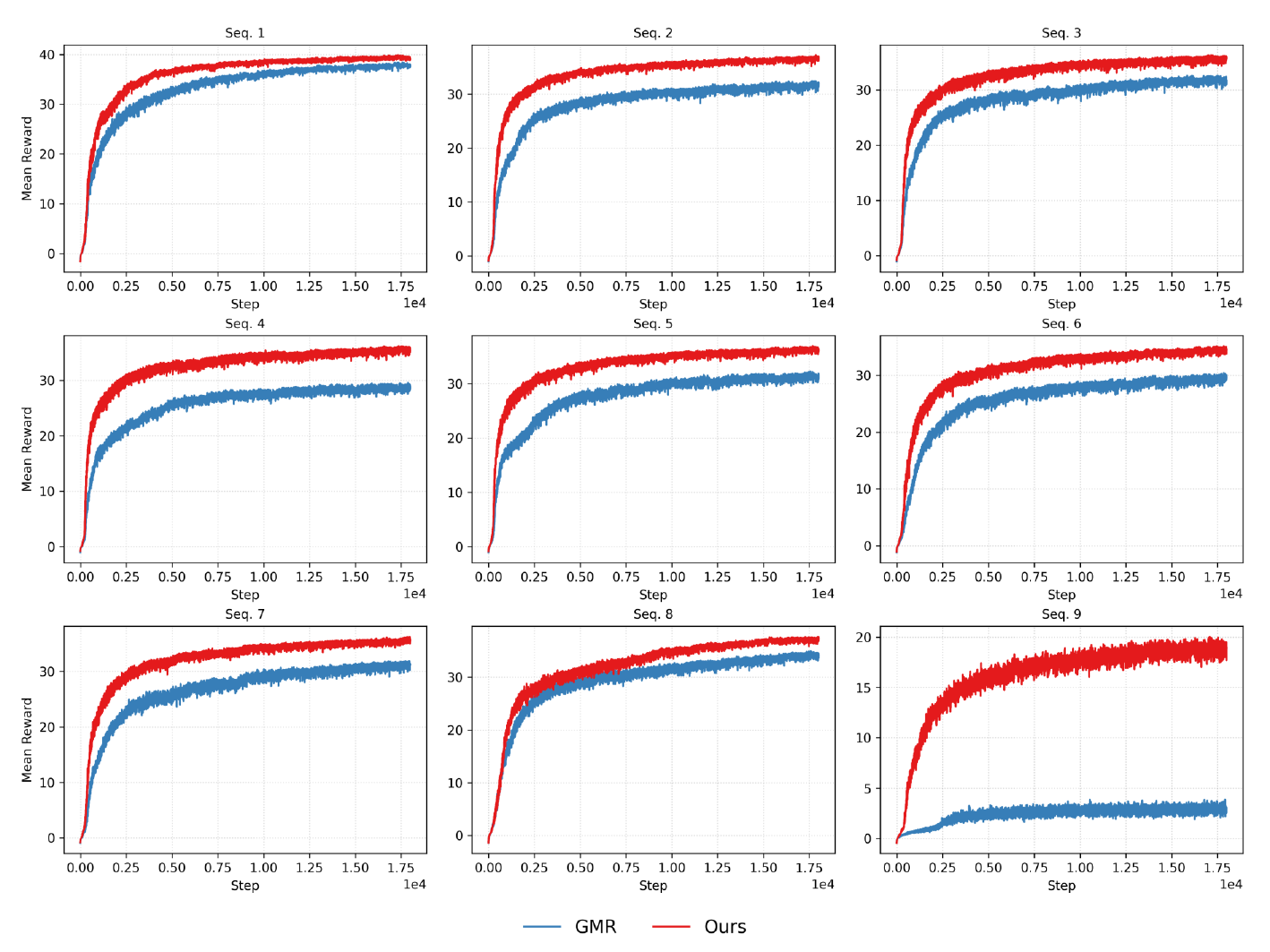}
\vspace{-4mm}
\caption{
The learning curves of mean reward for the Beyondmimic policy trained on GMR-retargeted and our retargeted motion.
}
\label{fig:beyondmimic_curve_mean_reward}
\vspace{-4mm}
\end{figure}

To better understand the exploration behavior, we logged the actions during the early stage of training (around 125 million samples) on Sideflip in single-trajectory motion tracking task. Figure~\ref{fig:joint_map} visualizes the resulting action distributions as heatmaps. The learning curves for the baselines and our method on the single-trajectory motion tracking, style-based motion mimicry, and general motion tracking tasks are shown in Figure~\ref{fig:ADD_singleclip_curve}, Figure~\ref{fig:AMP_singleclip_curve}, and Figure~\ref{fig:general_12}, respectively. Across all motions, our method converges consistently faster than the baselines in both tasks, indicating that PDF-HR substantially improves sample efficiency. Figure~\ref{fig:amass_RL_curve} further reports the learning curves of ADD~\cite{zhang2025physics} trained with GMR~\cite{araujo2025retargeting}-retargeted data versus our retargeted data, highlighting the superior quality of our retargeting. Finally, Figure~\ref{fig:AMP_single_clip} visualizes intermediate qualitative results on style-based motion mimicry after 350M training samples for all policies.

\begin{figure}[!ht] 
\centering
\includegraphics[width=0.99\linewidth]{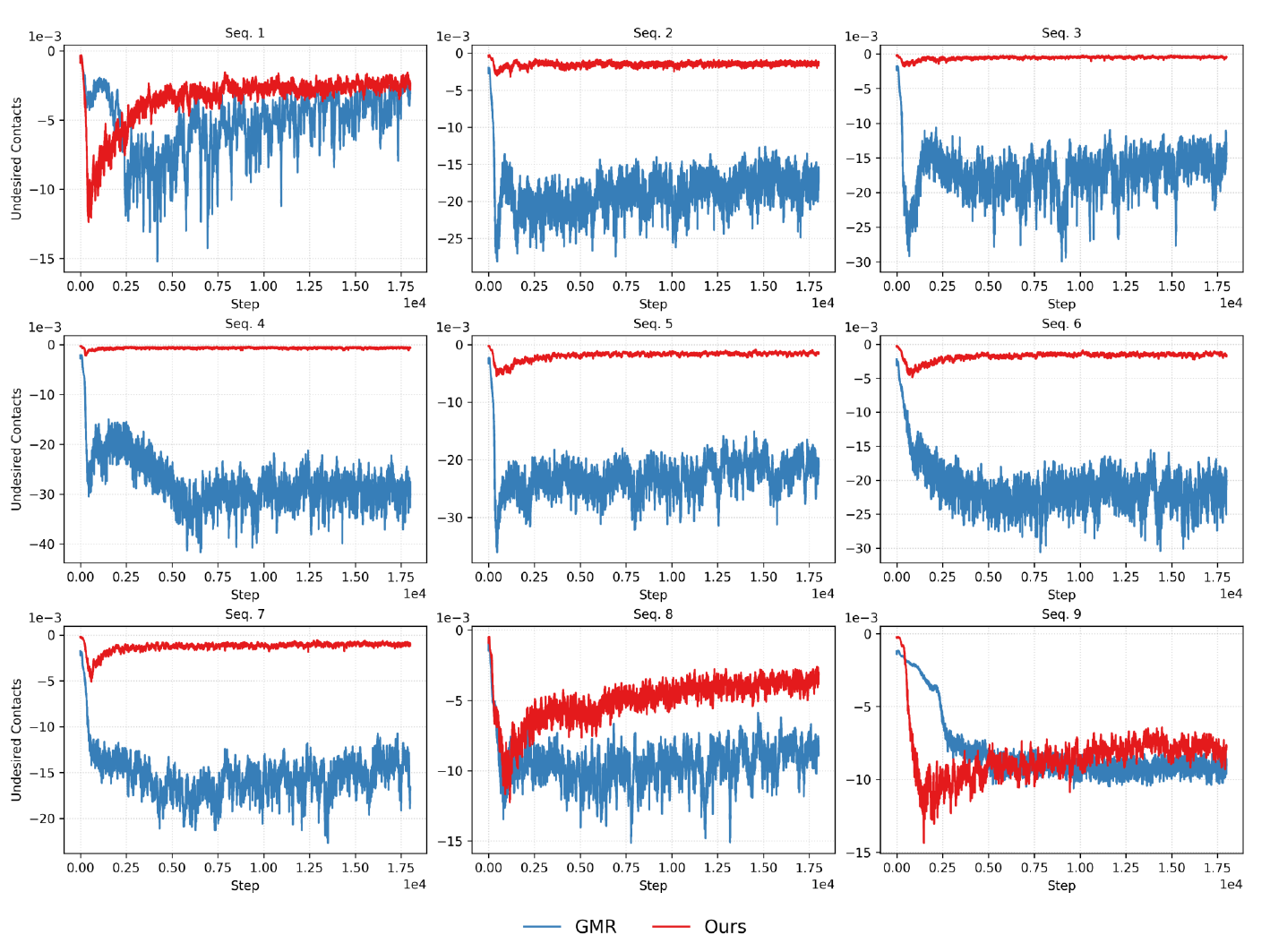}
\vspace{-3mm}
\caption{
The learning curves of undesired contacts for the Beyondmimic policy trained on GMR-retargeted and our retargeted motion.
}
\label{fig:beyondmimic_curve_undesired_contacts}
\vspace{-3mm}
\end{figure}
\begin{figure*}[!ht] 
\centering
\includegraphics[width=0.99\linewidth]{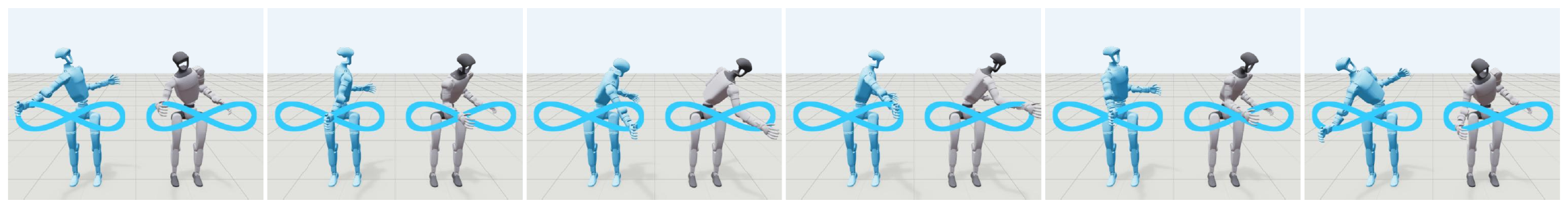}
\caption{\textbf{Visual comparisons of classical IK and our HL-IK.} We set the right-wrist end-effector to track an $\infty$-shaped trajectory. Classical QP-based IK (blue, left) meets the EE targets but can yield less natural whole-arm configurations, whereas our PDF-HR-regularized HL-IK (gray, right) produces more human-like arm postures while tracking the same trajectory.
}
\label{fig:hlik} 
\end{figure*}
Figure~\ref{fig:beyondmimic_curve_joint_pos_error}, Figure~\ref{fig:beyondmimic_curve_mean_reward}, and Figure~\ref{fig:beyondmimic_curve_undesired_contacts} report the learning curves of joint position error, mean reward, and undesired contacts for the Beyondmimic~\cite{liao2025beyondmimic} policy trained on GMR-retargeted motion and our retargeted motion. Under the same training protocol as defined in Beyondmimic, policies trained with our retargeted motions consistently converge faster and reach better plateaus across the nine sequences: the joint position error in Figure~\ref{fig:beyondmimic_curve_joint_pos_error} drops rapidly to a much lower level, while the GMR-retargeted counterpart typically stalls at a noticeably higher tracking error. This improved tracking quality is reflected in Figure~\ref{fig:beyondmimic_curve_mean_reward}, where our policy achieves higher mean rewards, indicating that the controller can satisfy task objectives more effectively once trained on physically consistent demonstrations. Moreover, Figure~\ref{fig:beyondmimic_curve_undesired_contacts} shows that our retargeted data substantially reduces the undesired-contact penalty (often by a large margin) and yields smoother, less noisy curves, suggesting more stable learning dynamics and fewer spurious contacts during execution. Overall, our retargeting gives the policy better training data: the motions stay within kinematic limits and the contact timing is more consistent, which leads to better training and easier deployment.

\section{Other Applications}
\label{appendix:extra_experiments}
\begin{figure}[!ht] 
\centering
\includegraphics[width=0.8\linewidth]{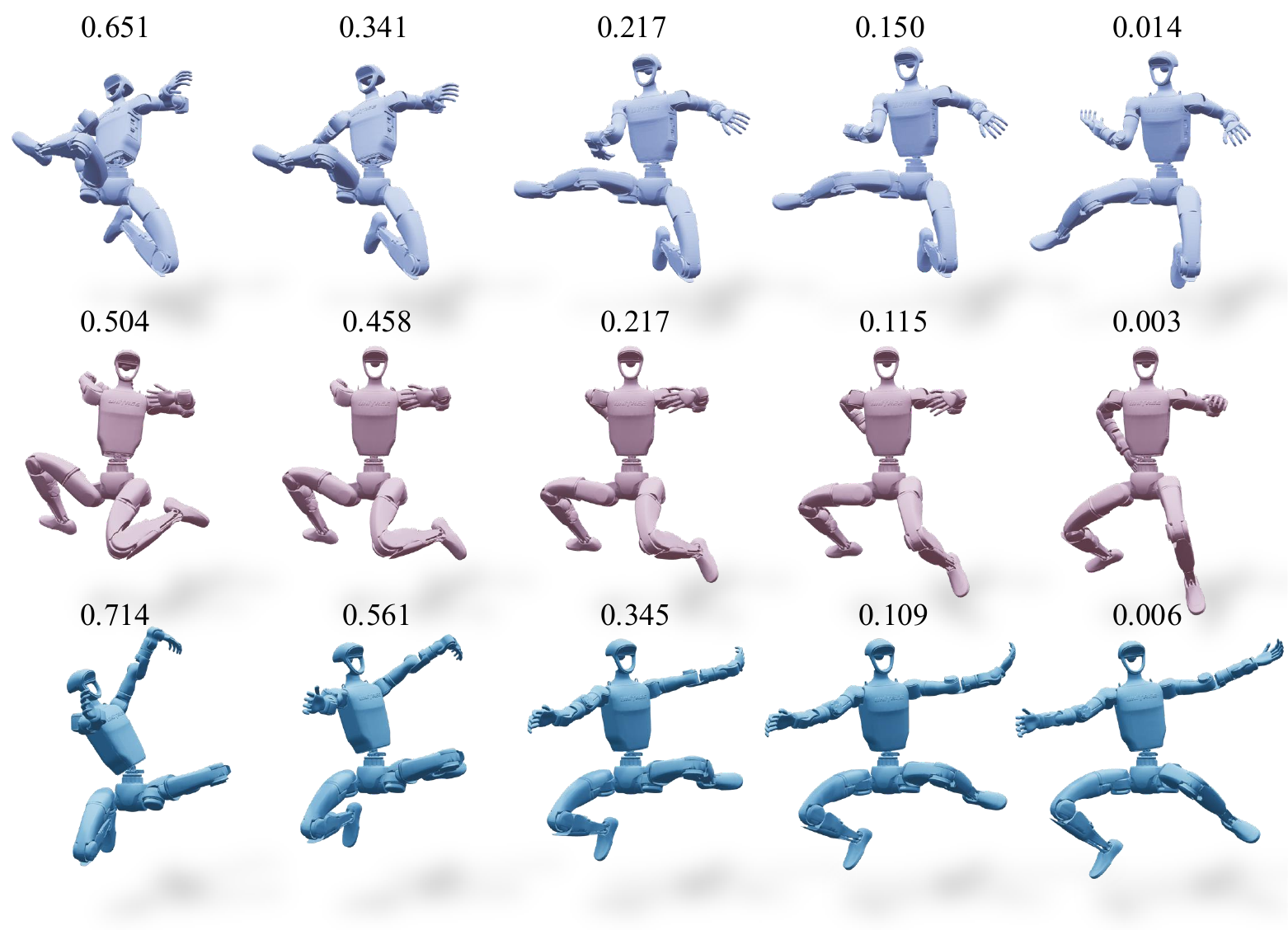}
\caption{
Visualization of the pose denoising process.
}
\label{fig:denoising} 
\end{figure}

\noindent
\textbf{Pose Denoising.} A critical application of our learned prior is pose denoising, which seeks to reconstruct physically plausible robot configurations from corrupted inputs. We leverage the pre-trained pose distance field as a fixed differentiable function to guide the optimization. Formally, the objective is to project a noisy query pose $\mathbf{q}_{\text{init}}$ onto the nearest point on the valid motion manifold. For evaluation, $\mathbf{q}_{\text{init}}$ is initialized via uniform sampling within joint limits. The pose is then iteratively refined by following the negative gradient of the distance field (Eq.~\eqref{eq:gradient}).
As qualitatively demonstrated in Figure~\ref{fig:denoising}, the optimization effectively rectifies kinematic violations. The trajectory evolves from an initial state exhibiting severe self-penetration and unnatural artifacts (left) to a collision-free, plausible configuration (right).

\noindent

\noindent
\textbf{Human-Like Inverse Kinematics (HL-IK).}
HL-IK~\cite{chen2025hl} introduces a \emph{human-like inverse kinematics} setting: given only end-effector (EE) targets at runtime, the solver should track the EE while producing whole-arm configurations that remain human-like, without relying on full-body sensing. Unlike HL-IK~\cite{chen2025hl}, which explicitly predicts non-EE joint/landmark targets, our method supports this task \emph{implicitly} by regularizing IK updates with the learned PDF-HR prior, making the approach more scalable and lightweight.

Since the official HL-IK~\cite{chen2025hl} implementation is not publicly available, we provide a qualitative demonstration. We define the right-wrist EE to follow a planar $\infty$-shaped trajectory and compare a classical QP-based IK solver against our PDF-HR-regularized variant, as shown in Figure~\ref{fig:hlik}. Concretely, we integrate PDF-HR into a quadratic-programming IK solver for the Unitree G1 upper body. At each control step, IK solves for a joint increment $\Delta\mathbf{q}$ that satisfies task-space objectives while remaining close to the learned motion manifold. Given the current configuration $\mathbf{q}_t$, we solve:
\begin{equation}
\begin{aligned}
\min_{\Delta \mathbf{q}} \;\;
& \tfrac{1}{2} w_{\text{task}}
  \big\| J(\mathbf{q}_t)\Delta \mathbf{q} + \mathbf{r} \big\|_2^2
+ \tfrac{1}{2}\lambda_{\text{smooth}}
  \big\| \Delta \boldsymbol{\theta} \big\|_2^2 \\
&\quad + \tfrac{1}{2}\lambda_{\text{PDF-HR}}
  \big\| \nabla f_\phi(\mathbf{q}_t)^\top \Delta \mathbf{q} \big\|_2^2,
\end{aligned}
\end{equation}
optionally subject to joint position and velocity limits. Here, $J$ and $\mathbf{r}$ denote the task Jacobian and residual (e.g., end-effector pose error). The PDF-HR term biases the update direction using the local normal of the learned manifold, $\nabla f_\phi(\mathbf{q}_t)$: by penalizing motion along the normal, the optimizer favors steps within the tangent space, yielding kinematically plausible and human-like joint updates while still meeting the same task-space targets. This soft regularization avoids explicit projection or hard feasibility constraints, yet keeps the solution close to the distribution of demonstrated human-like motions.

\noindent
\textbf{General Motion Tracking with PDF-HR.} To test whether PDF-HR generalizes to off-the-shelf motion trackers, we integrate our pose-prior reward into the low-level tracking policy of TextOp~\cite{TextOp_2025}. TextOp is a real-time, interactive framework for text-driven humanoid motion generation and control, featuring a two-layer design: a high-level text-conditioned diffusion autoregressive model generates a kinematic reference trajectory from the current user command, while a low-level universal tracking policy executes the trajectory on the robot to achieve responsive and accurate control.

\begin{figure}[!ht]
    \centering
    \begin{subfigure}[b]{0.49\linewidth}
        \centering
        \includegraphics[width=\linewidth]{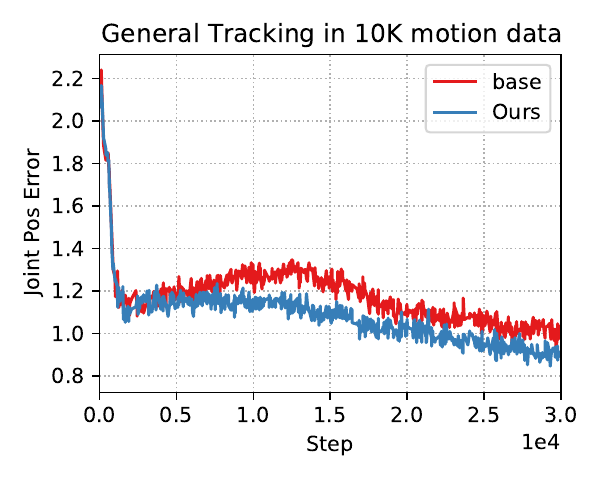}
        \vspace{-0.2in}
        \caption{Joint pose MSE per step} 
        \label{fig:gmt_plot1}
    \end{subfigure}
    \hfill
    \begin{subfigure}[b]{0.49\linewidth}
        \centering
        \includegraphics[width=\linewidth]{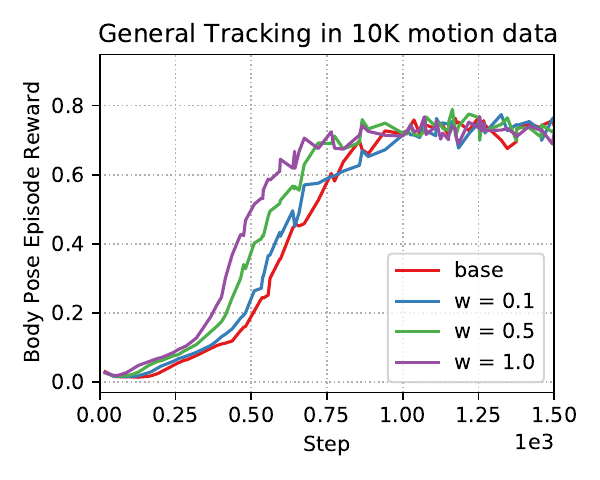}
        \vspace{-0.2in}
        \caption{Body pose episode reward} 
        \label{fig:gmt_plot2}
    \end{subfigure}
    \caption{The key learning curves of training general tracking controller with original reward and our pose-prior integrated reward in 10K motion data.}
    \label{fig:intro_main}
    
\end{figure}

We adopt TextOp's released low-level tracking policy as our baseline and follow the same training setup. Specifically, we train a general tracking policy on 10K motion clips sampled from AMASS~\cite{AMASS:ICCV:2019} and LaFAN1~\cite{harvey2020robust}. The only change from TextOp is that we additionally include the PDF-HR pose-prior reward (Eqs.~\eqref{eq:pose-prior score} and~\eqref{eq:pose-prior reward}) during training, while keeping the network, observations, and all other reward terms unchanged. As shown in Figure~\ref{fig:gmt_plot1}, our method consistently achieves lower joint pose MSE than the baseline throughout training. We further sweep the pose-prior weight $w$ to study its effect on the convergence rate. Figure~\ref{fig:gmt_plot2} indicates that increasing $w$ noticeably accelerates convergence: the body-pose episode reward rises earlier and reaches a high plateau in fewer steps, while the final performance remains comparable across settings. Overall, these results suggest that PDF-HR serves as a plug-and-play regularizer that improves both sample efficiency and tracking accuracy for general motion tracking.

\begin{figure}[!ht] 
\centering
\includegraphics[width=0.99\linewidth]{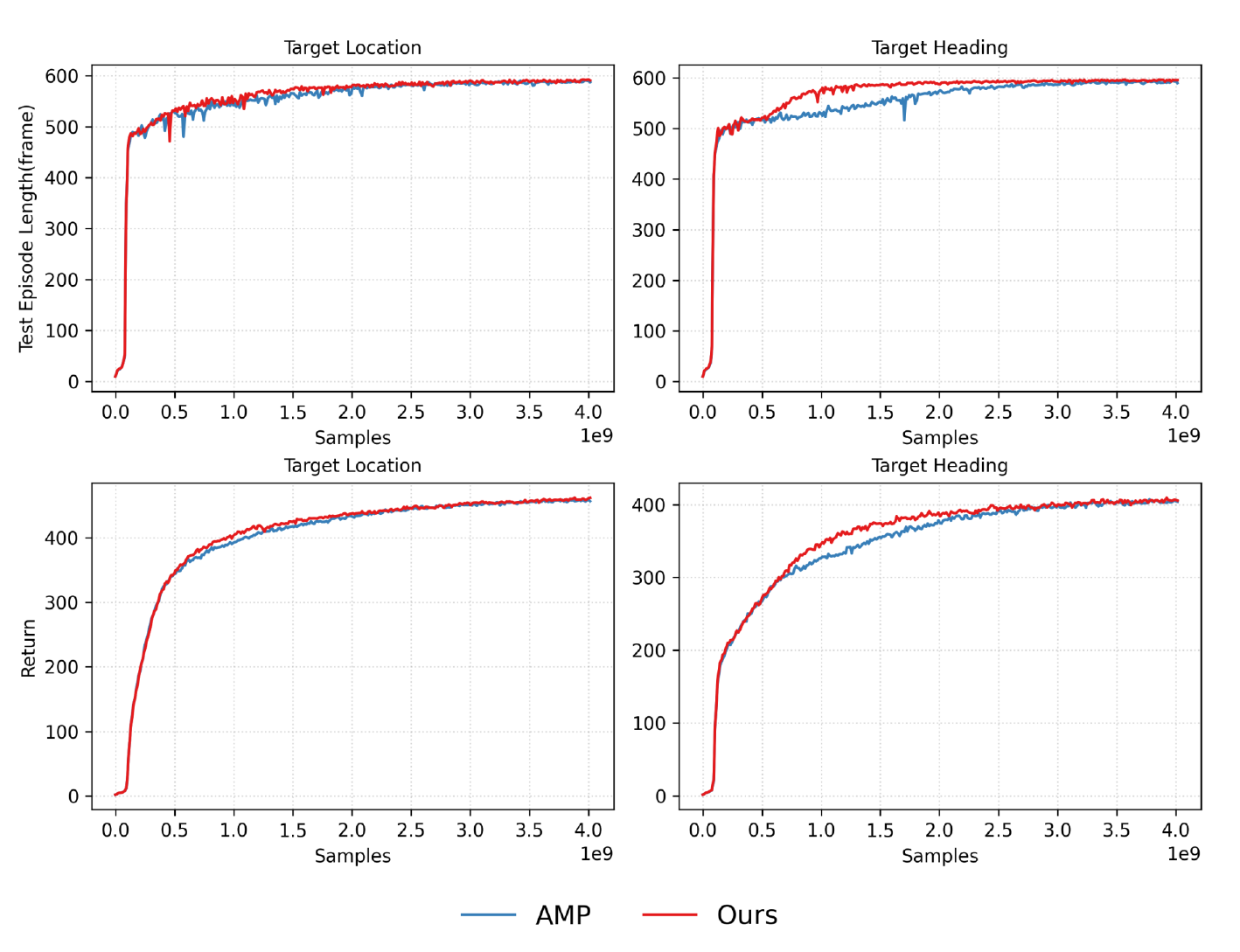}
\vspace{-4mm}
\caption{
The learning curves of AMP and our method on goal-conditioned tasks.
}
\label{fig:AMP_task_curve}
\end{figure}

\noindent
\textbf{Goal-conditioned Task.}
We further validated PDF-HR on two goal-conditioned tasks defined in AMP~\cite{peng2021amp}: Target Location (navigating to a random target position) and Target Heading (moving along a randomly changing direction at variable speeds). By incorporating our pose-prior reward into the AMP training objective with a weight $w^P=0.1$ and scale $\alpha^P=1.0$, we evaluated the test episode length and task return. As illustrated in Figure~\ref{fig:AMP_task_curve}, our method accelerates the convergence of the test episode length. This indicates that PDF-HR guides the agent to master basic locomotion skills faster. Consequently, this leads to better completion of high-level goals, resulting in higher task returns compared to the original AMP.

\end{document}